%% file: main.tex
\documentclass[twocolumn,letterpaper]{IEEEAerospaceCLS}  

\usepackage[]{graphicx}    
\usepackage{svg}
\newcommand{\ignore}[1]{}  
\usepackage{subcaption}

\usepackage{listings}
\usepackage{color}
\usepackage{float}
\usepackage{cite}
\usepackage{url}
\usepackage{tabularx}
\usepackage{hyperref}
\usepackage{tikz}

\definecolor{dkgreen}{rgb}{0,0.6,0}
\definecolor{gray}{rgb}{0.5,0.5,0.5}
\definecolor{mauve}{rgb}{0.58,0,0.82}

\newcommand\copyrighttext{%
  \footnotesize \textcopyright 2021 IEEE. Personal use of this material is permitted.
  Permission from IEEE must be obtained for all other uses, in any current or future
  media, including reprinting/republishing this material for advertising or promotional
  purposes, creating new collective works, for resale or redistribution to servers or
  lists, or reuse of any copyrighted component of this work in other works.}
\newcommand\copyrightnotice{%
\begin{tikzpicture}[remember picture,overlay]
\node[anchor=south,yshift=10pt] at (current page.south) {\fbox{\parbox{\dimexpr\textwidth-\fboxsep-\fboxrule\relax}{\copyrighttext}}};
\end{tikzpicture}%
}

\begin{document}
\title{A Pipeline for Vision-Based On-Orbit Proximity Operations Using Deep Learning and Synthetic Imagery}

\author{%
Carson Schubert, Kevin Black, Daniel Fonseka, Abhimanyu Dhir, Jacob Deutsch, Nihal Dhamani, Gavin Martin\\ 
Texas Spacecraft Laboratory\\
The University of Texas at Austin\\
Austin, TX 78712\\
carson.schubert14@gmail.com\\
\and 
Maruthi Akella\\
Department of Aerospace Engineering and Engineering Mechanics\\
The University of Texas at Austin\\
Austin, TX 78712\\
makella@mail.utexas.edu
\thanks{\footnotesize 978-1-7281-7436-5/21/$\$31.00$ \copyright2021 IEEE}              
}

\maketitle
\copyrightnotice

\thispagestyle{plain}
\pagestyle{plain}

\copyrightnotice

\maketitle

\thispagestyle{plain}
\pagestyle{plain}

\begin{abstract}
Deep learning has become the gold standard for image processing over the past decade. Simultaneously, we have seen growing interest in orbital activities such as satellite servicing and debris removal that depend on proximity operations between spacecraft. However, two key challenges currently pose a major barrier to the use of deep learning for vision-based on-orbit proximity operations. Firstly, efficient implementation of these techniques relies on an effective system for model development that streamlines data curation, training, and evaluation. Secondly, a scarcity of labeled training data (images of a target spacecraft) hinders creation of robust deep learning models. This paper presents an open-source deep learning pipeline, developed specifically for on-orbit visual navigation applications, that addresses these challenges. The core of our work consists of two custom software tools built on top of a cloud architecture that interconnects all stages of the model development process. The first tool leverages Blender, an open-source 3D graphics toolset, to generate labeled synthetic training data with configurable model poses (positions and orientations), lighting conditions, backgrounds, and commonly observed in-space image aberrations. The second tool is a plugin-based framework for effective dataset curation and model training; it provides common functionality like metadata generation and remote storage access to all projects while giving complete independence to project-specific code. Time-consuming, graphics-intensive processes such as synthetic image generation and model training run on cloud-based computational resources which scale to any scope and budget and allow development of even the largest datasets and models from any machine. The presented system has been used in the Texas Spacecraft Laboratory with marked benefits in development speed and quality. Remote development, scalable compute, and automatic organization of data and artifacts have dramatically decreased iteration time while increasing reproducibility and system comprehension. Diverse, high-fidelity synthetic images that more closely replicate the real environment have improved model performance against real-world data. These results demonstrate that the presented pipeline offers tangible benefits to the application of deep learning for vision-based on-orbit proximity operations.
\end{abstract}

\tableofcontents

\section{Introduction}
\label{sec:introduction}
\input{introduction.tex}

\section{Motivation}
\label{sec:motivation}
\input{motivation.tex}

\section{Existing Tools}
\label{sec:existing_tools}
\input{existing_tools.tex}

\section{Synthetic Image Generation}
\label{sec:image_gen}
\input{synthetic.tex}

\section{Pipeline Overview}
\label{sec:pipeline}
\input{pipeline.tex}

\section{Case Examples}
\label{sec:cases}
\input{case_examples.tex}

\section{Conclusion}
\label{sec:conclusion}
\input{conclusion.tex}






\acknowledgments
We would like to thank Pratyush Singh and Christian Sweet for their contributions at the onset of pipeline development, and Evan Wilde for his support with Blender along the way. We also thank Siddarth Kaki for his support on a variety of research topics. Finally, we thank the NASA Johsnon Space Center, in particular Sam Pedrotty, which provided funding that contributed to this work. 

Trademark notice: Amazon Web Services, S3, the S3 logo, EC2, the EC2 logo, and SageMaker are trademarks of Amazon.com, Inc. or its affiliates in the United States and/or other countries.

\bibliographystyle{IEEEtran}
\bibliography{IEEEabrv, references.bib}





\thebiography
\vspace{-4 mm}
\begin{biographywithpic}
{Carson Schubert}{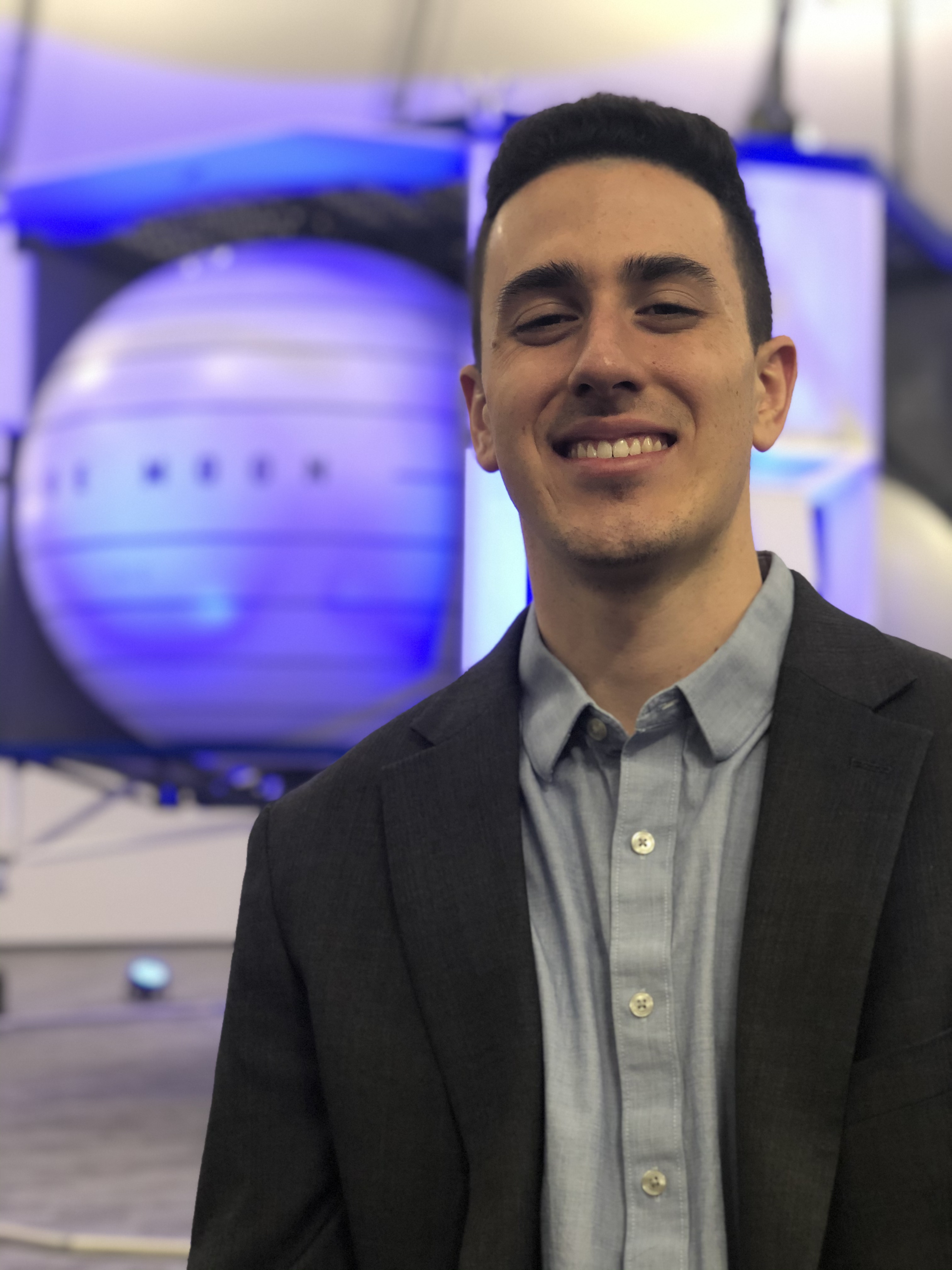}
is currently enrolled as an undergraduate student in both the Cockrell School of Engineering and the College of Natural Sciences at the University of Texas at Austin studying electrical engineering and mathematics. His academic focus is on wireless communications, signal processing, networks, and systems. Carson has worked as an undergraduate researcher in the Texas Spacecraft Laboratory since 2017.
\end{biographywithpic} 
\begin{biographywithpic}
{Kevin Black}{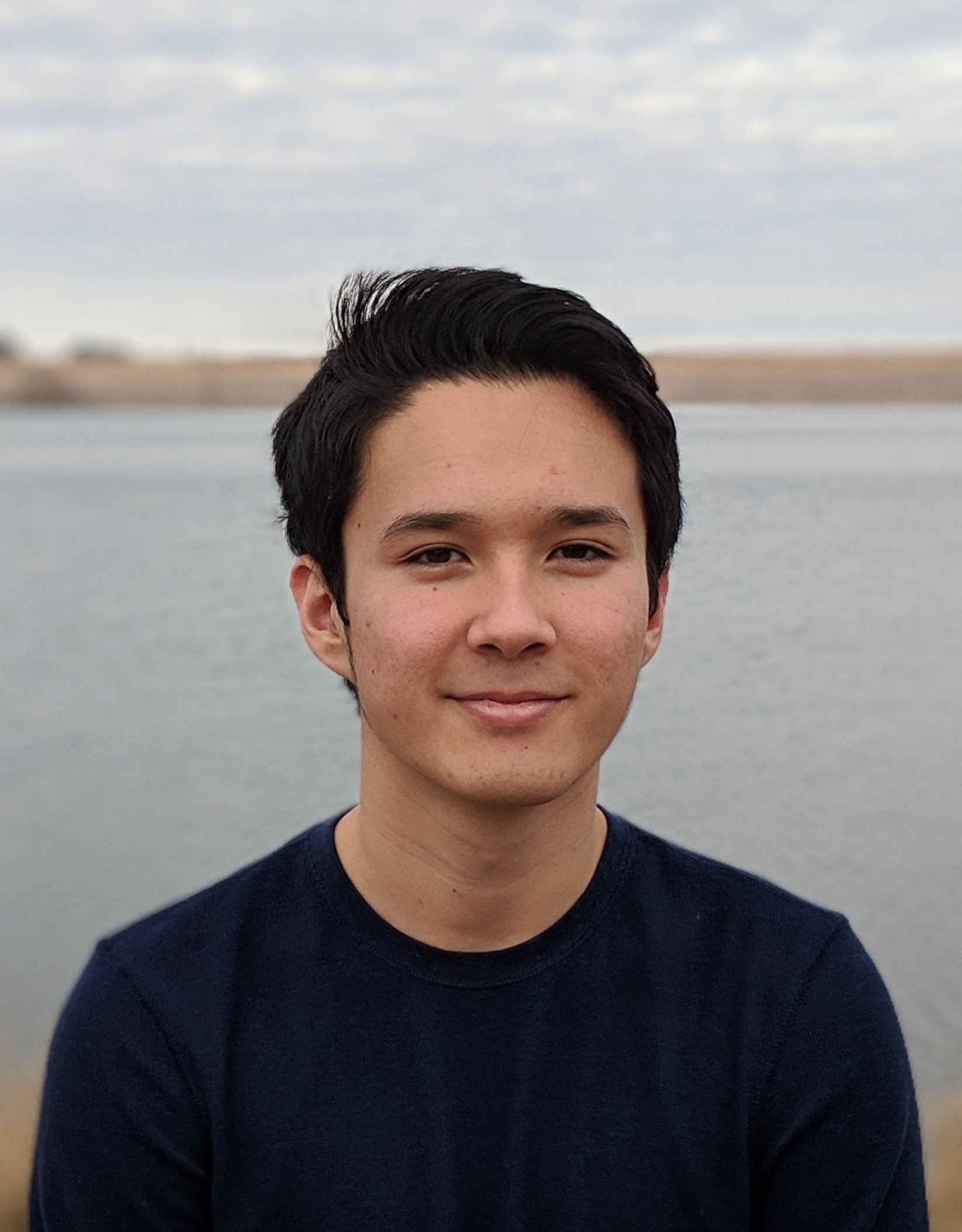}
is currently enrolled as an undergraduate student in the College of Natural Sciences at the University of Texas at Austin studying computer science and mathematics. He is in the Turing Scholar's and Dean's Scholars honors programs, with an academic focus on computer vision and machine learning. Kevin has worked as an undergraduate researcher in the Texas Spacecraft Laboratory since January 2019.
\end{biographywithpic} 
\begin{biographywithpic}
{Daniel Fonseka}{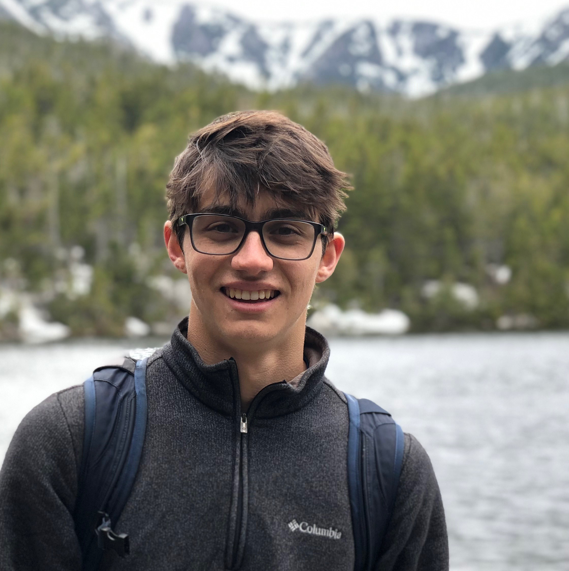}
is currently enrolled as an undergraduate student in the Cockrell School of Engineering at the University of Texas at Austin studying Computational Engineering. His academic focus is on data science and software engineering. Daniel has worked as an undergraduate researcher in the Texas Spacecraft Laboratory since January 2019.
\end{biographywithpic} 
\begin{biographywithpic}
{Abhi Dhir}{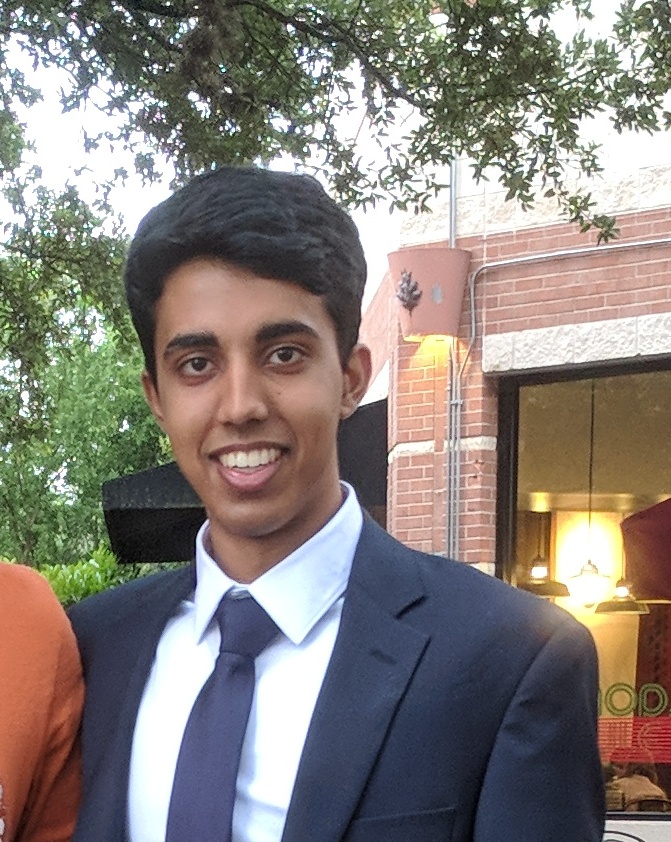}
is currently enrolled as an undergraduate student in the College of Natural Sciences at the University of Texas at Austin studying computer science and mathematics. His academic focus is on software engineering and machine learning. Abhi has worked as an undergraduate researcher in the Texas Spacecraft Laboratory since March of 2020.
\end{biographywithpic} 
\begin{biographywithpic}
{Jacob Deutsch}{Figures/jacob_face.JPG}
is currently enrolled as an undergraduate student in the Cockrell School of Engineeering at the University of Texas at Austin studying Electrical and Computer Engineering. His academic focus is Data Science and Software Engineering. Jacob has worked as an undergraduate researcher in the Texas Spacecraft Laboratory since March of 2020. In his free time, Jacob enjoys baking cookies and doing personal data science projects.
\end{biographywithpic} 

\begin{biographywithpic}
{Nihal Dhamani}{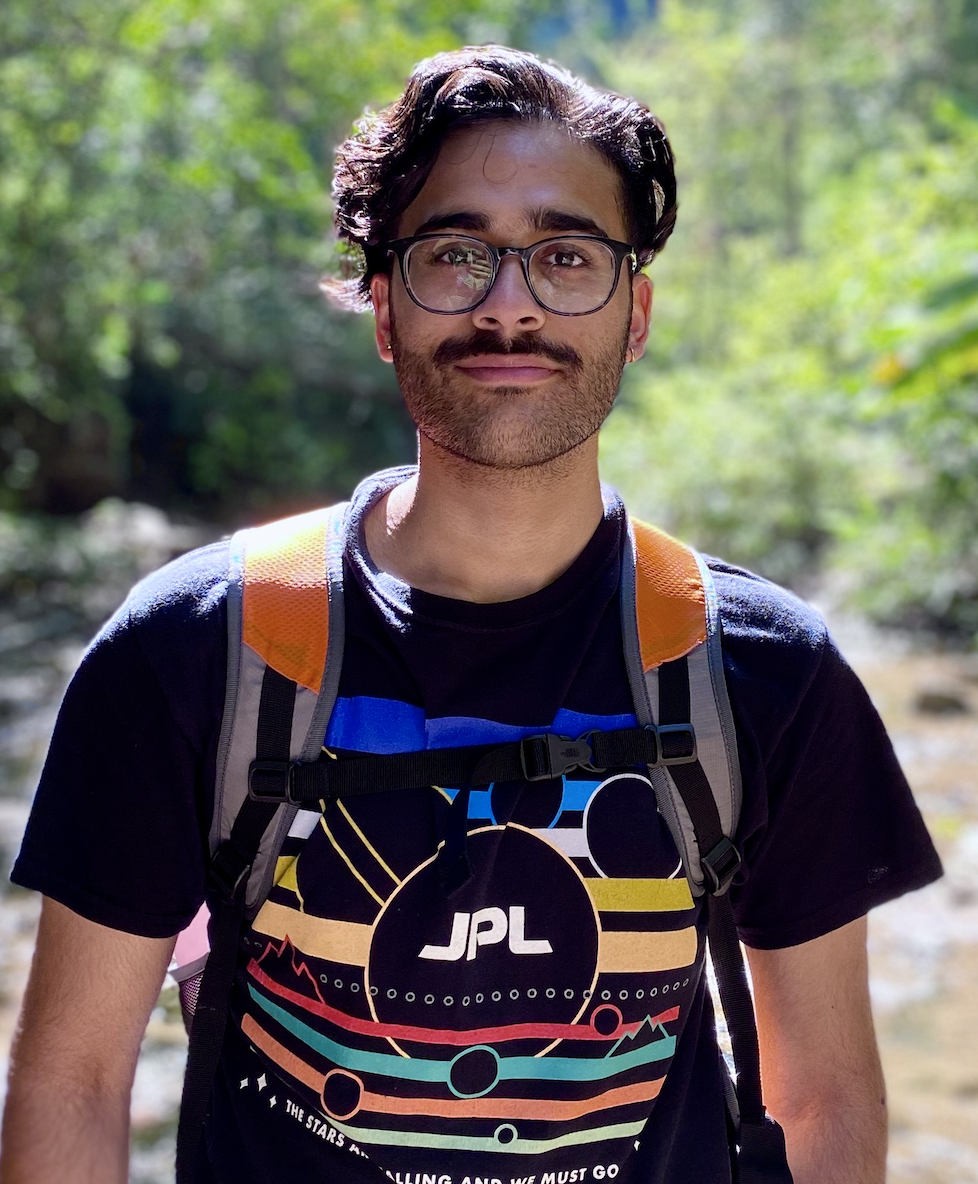}
Nihal Dhamani graduated from the College of Natural Sciences at the University of Texas at Austin in December 2019, majoring in both Computer Science and Astronomy. His academic focus was in Machine Learning and Artificial Intelligence and he worked as an undergraduate researcher at the Texas Spacecraft Laboratory from 2017-2019. Currently, he works at NASA Jet Propulsion Laboratory.
\end{biographywithpic}

\begin{biographywithpic}
{Gavin Martin}{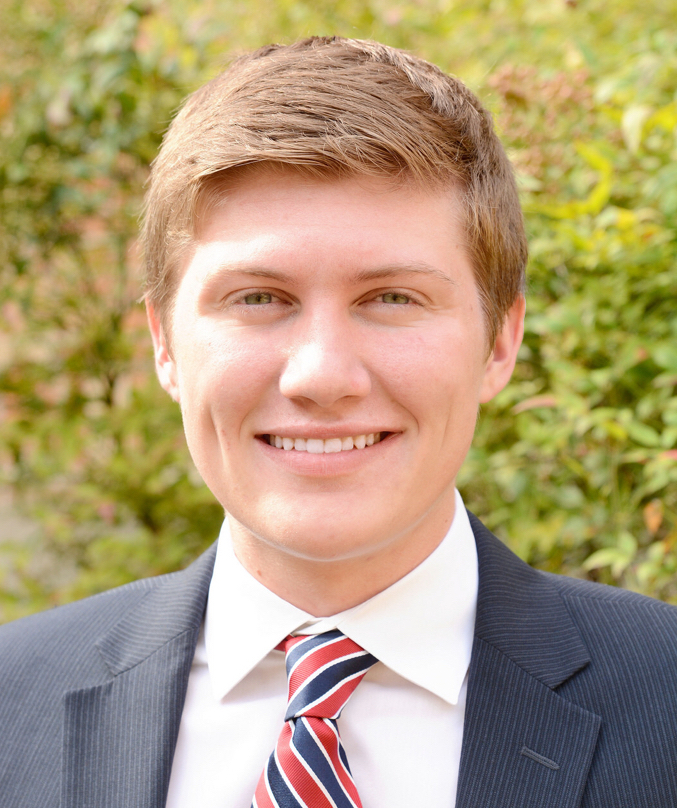}
Gavin Martin was an undergraduate Aerospace Engineering student at the University of Texas at Austin while this research was conducted. He was a researcher in the Texas Spacecraft Laboratory at the time, and his interests span autonomous systems, modeling and simulation, and small satellites. He has since graduated from UT-Austin and is a Software Systems Engineer at NASA’s Jet Propulsion Laboratory, where he works on planning and execution software for the Europa Clipper mission.
\end{biographywithpic}

\begin{biographywithpic}
{Maruthi R. Akella}{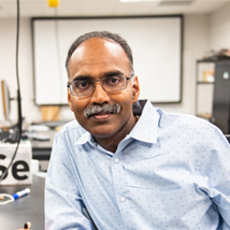}
is a tenured faculty member with the Department of Aerospace Engineering and Engineering Mechanics at The University of Texas at Austin (UT Austin) where he holds the Ashley H. Priddy Centennial Professorship in Engineering. He is the founding director for the Center for Autonomous Air Mobility and the faculty lead for the Control, Autonomy, and Robotics area at UT Austin. Dr. Akella’s research program encompasses control theoretic investigations of nonlinear and coordinated systems, vision-based sensing for state estimation, and development of integrated human and autonomous multivehicle systems. He is Editor-in-Chief for the Journal of the Astronautical Sciences and Technical Editor (Space Systems) for the IEEE Transactions on Aerospace and Electronic Systems. Dr. Akella is a Fellow of the AAS, a Senior Member of the IEEE, and serves as IEEE Distinguished Lecturer for the Aerospace and Electronic Systems Society.
\end{biographywithpic}

\end{document}

%% file: introduction.tex
The superiority of deep learning techniques for image processing has been proven over the past decade. Accordingly, we have seen a growing body of work like \cite{Cassinis2020, Park2019, Proenca2019, Sharma2019} that applies these techniques to the problem of vision-based on-orbit proximity operations. Vision-based proximity operations are of particular interest over other approaches due to the low size, weight, and power requirements associated with visual sensors such as monocular cameras; as such, they are suitable for a broad class of spacecraft. Additionally, these sensors are low cost and widely available from commercial vendors. While results are promising, the application of deep learning for this purpose is currently plagued by two major challenges that stifle progress and hinder system performance.

The first of these challenges relates to the model development process. In the context of deep learning, a model is the input-output relationship one trains to perform a function such as classification or feature extraction. Model development generally consists of three high-level tasks: data curation, training, and evaluation. In a standard model development process, including that used for models targeting on-orbit proximity operations, these tasks will often be executed dozens or hundreds of times as developers experiment with varying datasets, model architectures, and training configurations. Such experimentation will produce an equally vast array of trained models and performance metrics that must be stored, tracked, and compared against one another. Clearly, an effective system for running these tasks and organizing their outputs is a boon to development efficiency and success. In the absence of such a streamlined system, project codebases often become a complex web of undocumented datasets, training artifacts, and ad-hoc code, hindering development speed and leading to irreproducible results. This challenge is common to all deep learning applications regardless of domain; however, it is particularly relevant in the space domain due to strict reliability standards that necessitate enhanced system comprehension.

The second challenge is a scarcity of labeled on-orbit images. Deep learning models require a large, diverse set of labeled images for successful training, even when techniques such as transfer learning are employed. For the purpose of proximity operations, these images are generally of a target object or spacecraft in a variety of orbital scenes. Unfortunately, it is infeasible to obtain a sufficiently diverse dataset of real images for the vast majority of objects of interest, and for spacecraft that have not yet flown, it is impossible. Furthermore, the images that are available are rarely labeled with ground truth data, such as vehicle orientation, rendering them nearly useless for deep learning purposes. While images can be hand labeled, this process is arduous and, in the case of complex labels like vehicle orientation, error-prone. For these reasons nearly all applications of deep learning for vision-based on-orbit proximity operations have utilized synthetic imagery, including \cite{Cassinis2020, Park2019, Proenca2019, Sharma2019}. The use of synthetic imagery for training deep learning models is an active area of research; it stands to eliminate the expensive task of hand labeling images and enable deep learning systems in new operating environments like low-earth-orbit (LEO). There are a number of promising results from the past five years like \cite{Richardson2016, Hinterstoisser2019, camera_pose, Tremblay2018}. However, it is uniquely difficult to simulate realistic orbital scenes due to the diversity of lighting conditions, vehicle orientations, and image noise present in the space environment. A major takeaway from our prior work on the NASA Seeker mission \cite{SeekerSciTech} is that models trained on a sub-par synthetic image dataset, both in terms of fidelity and diversity, can struggle to generalize well to the real environment, a finding in line with other works like \cite{Talwar2020}.

In this paper we present an open-source deep learning pipeline developed specifically to address the aforementioned challenges. It comprises two custom software tools that enable rapid, flexible, and repeatable development of deep learning models for vision-based on-orbit proximity operations. These novel tools integrate seamlessly with cloud resources, such as storage and compute, that allow the pipeline to scale to any scope, schedule, or budget. This pipeline has been used with great success by the Texas Spacecraft Laboratory on multiple research tasks over the past year.

We first detail our motivation for pipeline development by describing the difficulties faced during work on the Seeker Vision system. For context, we then compare the presented system to the myriad open-source and commercial tools available today for streamlining model development. The pipeline architecture and individual components are then described. Finally, we give four case examples of pipeline use and observed benefits, including a comparison of model performance before and after using our improved synthetic imagery.

%% file: motivation.tex
The Seeker Vision system, detailed in \cite{SeekerSciTech} and flown on NASA JSC's Seeker 1 Cubesat \cite{Seeker1}, utilized a convolutional neural network to identify and localize the Cygnus spacecraft in monocular imagery. Development and performance of this model was hindered by both challenges described previously: an inefficient dataset curation, training, and evaluation system along with sub-par synthetic imagery. Here we describe the tooling used for Seeker Vision to establish a comparative baseline and motivation for the work presented in this paper.

\begin{figure}[htpb]
    \includegraphics[width=\linewidth]{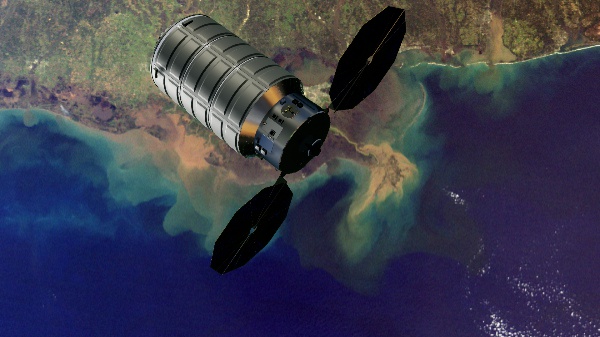}
    \caption{Synthetic Cygnus on synthetic Earth background.}
    \label{fig:synth_cygnus_synth_back}
\end{figure}

\begin{figure}[htpb]
    \includegraphics[width=\linewidth]{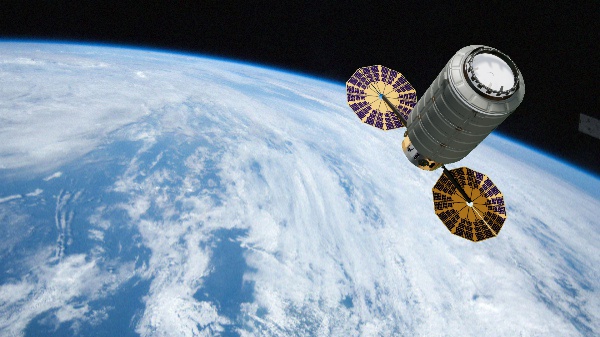}
    \caption{Synthetic Cygnus on real Earth background.}
    \label{fig:synth_cygnus_real_back}
\end{figure}

\begin{figure}[htpb]
    \includegraphics[width=\linewidth]{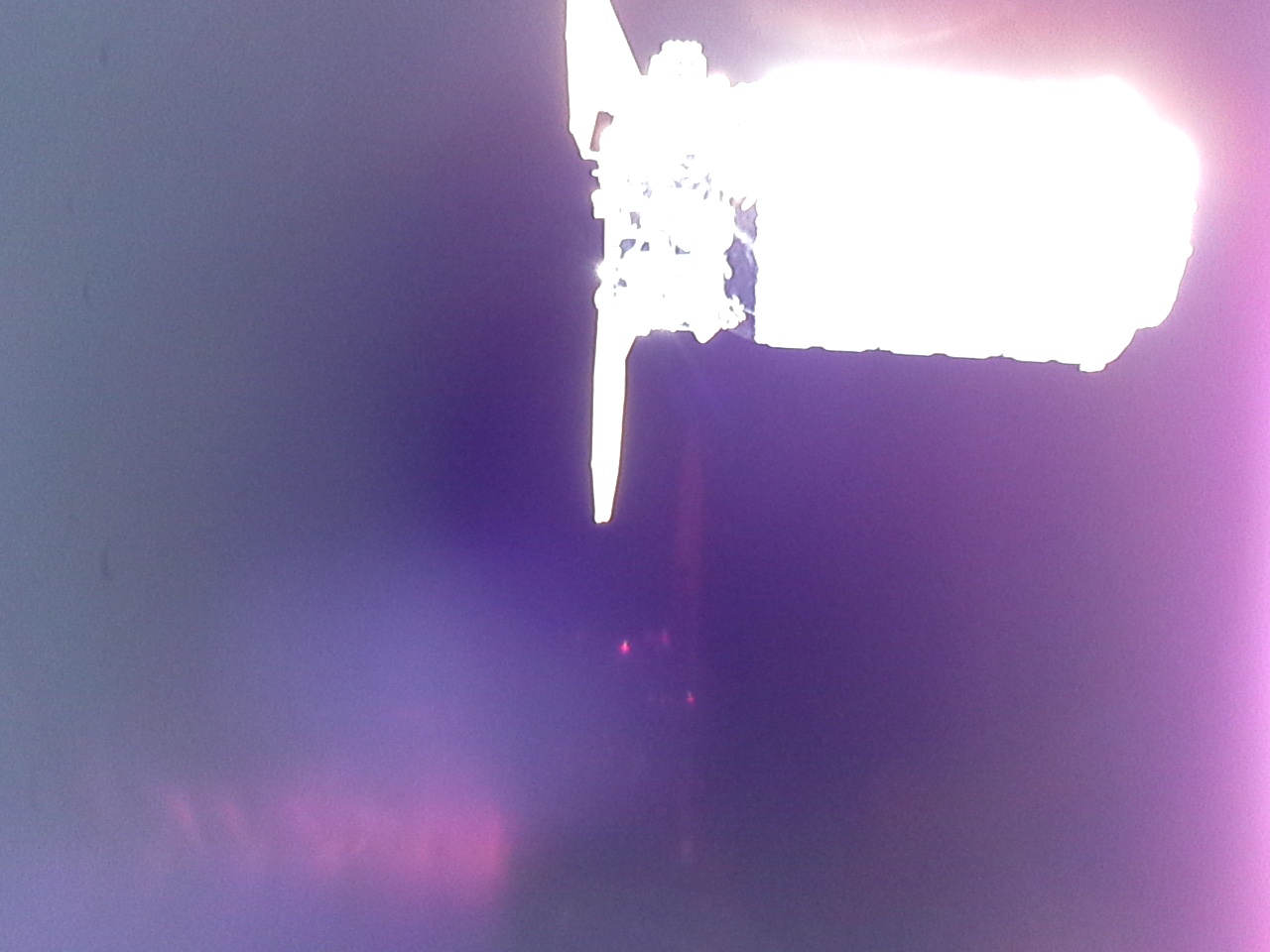}
    \caption{In-flight image from NASA JSC's Seeker 1 Mission \cite{Seeker1} captured by Seeker Vision.}
    \label{fig:skr1_inflight}
\end{figure}

The video game engine Unreal Engine 4 was used to generate a synthetic image dataset for training. Images feature an open-source Blender model of Cygnus \cite{skr1_blender_sc_models} on a variety of synthetic and real backgrounds. For synthetic backgrounds, the direct output of Unreal Engine was used; an example image is shown in Figure \ref{fig:synth_cygnus_synth_back}. For real backgrounds, Cygnus was rendered in front of a purely green background, then extracted with a mask and placed over real images of Earth and deep space from low-earth orbit (LEO); an example image is shown in Figure \ref{fig:synth_cygnus_real_back}. Image generation was controlled via Python scripts that rotate and translate the image capture point-of-view around the spacecraft to give a variety of relative poses. However, each generation run had static lighting and background settings that required manual alteration between runs. Thus, creation of diverse datasets was a multi-stage process gated by human interaction. While the produced synthetic data was of relatively high quality, lack of a physically-based, ray-tracing rendering engine limited realism, particularly with regard to lighting and texture. Post-flight evaluation revealed that the training data was not fully representative of the real operating environment, where aberrations such as light blooming and severe image blur occurred frequently, and we observed the model struggle in these cases. Figure \ref{fig:skr1_inflight} shows an example image from the mission that displays these aberrations.

All datasets and training artifacts (models and performance metrics) were manually organized and maintained, and no metadata such as date of creation, author, or code version was automatically associated with these objects. This made evaluation difficult and error-prone; if a user poorly named their trained model, it could easily become impossible to later determine which dataset and hyperparameters were used for training. Paradoxically, attempting to remedy this problem by enforcing a robust organization and naming scheme on developers had an equally negative effect; development efficiency and speed is simply traded for an improvement in rigor. Even with proper organization, model comparison was difficult without a graphical interface to automatically visualize performance metrics and track trained models. Both data generation and model training could only be conducted locally on a desktop computer with a single NVIDIA graphics processing unit (GPU); this constituted a serious bottleneck. Simultaneous model training or data generation was impossible, and hardware was fixed, thus placing a hard limit on runtimes without the purchase of additional or upgraded computational resources that are expensive and often take weeks to arrive. Here again an undesirable tradeoff emerged, this time between freedom for experimentation and development speed.

These experiences make it evident future research depends on significant improvements to both synthetic data and the processes used for model development. The tooling used for Seeker Vision, while successful at producing a flight-ready system, was not sustainable or efficient and would stunt progress toward the full six-degree-of-freedom relative pose estimation system we hoped to develop. More importantly, a deep learning-based system for on-orbit proximity operations that is tied to a single target spacecraft has significantly less utility. One-off solutions make widespread use impractical and financially infeasible; rather, the system must be capable of quick, simple application to almost any object of interest.

We therefore sought to create a model development pipeline that, in addition to addressing many of the inefficiencies described above, can be combined with a particular deep learning architecture to create a broadly applicable system which requires minimal tuning for each particular use case. The first stage of this pipeline is clearly the generation of higher quality and more diverse synthetic data; in particular, the problem of relative pose estimation demands this, as such estimation is far more sensitive to local lighting and texture features than simpler object detection and localization. Thus we also sought to develop a flexible, realistic image generation system that can quickly produce a quality dataset for any target object of interest. Crucially, we required that these tools be low-cost and, ideally, open-source, to minimize barrier to entry. As stated previously, a major goal is enabling the widespread adoption of deep learning systems for on-orbit proximity operations. A model development system based on expensive enterprise tools, while likely to be quite functional, is impractical for all but the most well-funded (or simply massive enough) organizations.

%% file: existing_tools.tex
Over the past five years, driven by massive growth in the popularity of deep learning, a host of platforms have emerged to assist with model development. These platforms aim to solve the problems we have described by managing certain boilerplate tasks like data storage and computational resource allocation automatically. Rather than roll their own solution, users merely configure these tasks for their use case through simple interfaces exposed by the platform. It should be noted that not all platforms attempt to address the entire model creation process. We limit our discussion exclusively to those that offer an end-to-end solution for brevity and direct comparison to the presented work.

As one might expect, the landscape of platforms is rapidly evolving in this nascent product space where competition is fierce and a single feature advantage can make a huge difference in user adoption. Platform offerings and pricing change significantly month over month. For simplicity, here we detail only features and pricing available at the time of writing; no attempt is made to give a historical or future perspective. The work presented in this paper was started in January of 2019; at that time, no available platform met our particular needs, especially given cost considerations. We argue the custom solution developed to meet that need and presented in this paper remains competitive today.

As stated previously, cost was a top concern. Some large companies with robust pre-existing cloud infrastructure charge users only for the cloud resources utilized by the platform's features. The features themselves, such as visualization dashboards, pipeline orchestration, and managed allocation of computational resources, are provided at no cost. Google's AI Platform \cite{googleAI} and Microsoft's Azure Machine Learning \cite{azureML} are two notable platforms which follow this model. In contrast, Amazon's SageMaker \cite{sagemaker} requires that training and optimization run on special ``ml'' compute instances. Averaged across GPU-enabled instances, these cost 30\% more per hour than the equivalent instance used outside of SageMaker on Amazon's cloud compute service EC2 \cite{ec2}, thus constituting a charge for SageMaker features. A few representative price comparisons for these instances are shown in Table \ref{tab:sagemaker_prices}. While there is a premium, cost is still solely dependent on actual usage.

\begin{table}[htpb]
    \centering
    \renewcommand*{\arraystretch}{1.4}
    \caption{Hourly price comparison between SageMaker ML instances and equivalent EC2 instances \cite{sagemaker_pricing, ec2_pricing}.}
    \label{tab:sagemaker_prices}
    \begin{tabularx}{\linewidth}{c|c|c|c|X}
        \hline
        Type & EC2 & SageMaker & Premium & Hardware \\
        \hline
        p3.2xl & \$3.06 & \$3.825 & 25\% & NVIDIA V100 x1 \\
        p3.8xl & \$12.24 & \$14.68 & 20\% & NVIDIA V100 x4 \\
        p2.16xl & \$14.40 & \$16.56 & 15\% & NVIDIA K80 x16 \\
        g4dn.xl & \$0.526 & \$0.736 & 40\% & NVIDIA T4 x1
    \end{tabularx}
\end{table}

Other platforms charge a monthly subscription fee on top of usage costs, such as Paperspace's Gradient \cite{paperspace_gradient}, allegro.ai enterprise \cite{allegro_enterprise}, and Polyaxon enterprise \cite{polyaxon_enterprise}. Gradient and Polyaxon (for more than three person teams) charge a fee per-user; pricing information for allegro.ai was not available at the time of writing. In the case of Gradient, this monthly fee is offset by access to cheaper compute instances. For comparison we consider instances with a single NVIDIA V100 GPU, one of the most powerful in the world. Compared to the equivalent raw Amazon EC2 instance, p3.2xl, Gradient is 25\% cheaper per hour, and compared to the SageMaker ``ml'' version it is 40\% cheaper per hour \cite{gradient_pricing}. While this is substantial, consider that a six person team using Gradient's lowest team plan T1, at \$12 per user, would incur \$72 in monthly subscription fees. In order to break even via hourly price savings, this team would need to already utilize over 94 hours of EC2 compute or 47 hours of SageMaker ``ml'' compute. Any subscription based model requires some level of minimum use to be practical. Active teams, such as those at technology firms who are training, deploying, and re-training models continually, may easily meet this threshold; however, using these platforms as a foundation for a model development pipeline targeted at the smaller, research-oriented field of on-orbit proximity operations is not practical.

Finally, there are the open-source platforms whose cost, like AI Platform and Azure Machine Learning, is solely dependent on actual cloud storage and compute usage. The key difference is they intentionally do not tie themselves to a specific cloud provider; rather, it is up to the user if and how to utilize cloud services. MLFlow \cite{MLFlow} and Kubeflow \cite{Kubeflow} are the most prominent and feature rich among these; they are entirely open-source, with no parallel enterprise offering. In contrast, Polyaxon and allegro.ai both offer an open-source version with core functionality \cite{allegro_open_source, polyaxon_ce}, but lock certain features behind the enterprise paywall discussed earlier .

Clearly, if cost is a top priority, platforms which do not charge a premium to access their features are desirable. We narrow our discussion of features to those platforms, thus excluding SageMaker and the enterprise offerings from Polyaxon and allegro.ai. Many features are common to all platforms, such as hyperparameter tuning, experiment tracking, artifact management, and support for popular deep learning frameworks. Differences emerge in three main areas. The first is flexibility in cloud services. AI Platform and Azure Machine Learning, as a condition of their free feature set, require use of those company's respective cloud storage and compute. None of the open-source platforms discussed above have this restriction, though Kubeflow and Polyaxon do require use of the open-source Kubernetes engine \cite{kubernetes}, a cluster-based cloud architecture. 

The second area of difference is allocation and management of cloud resources. All the open-source platforms mentioned (MLFlow, Polyaxon, Kubeflow, and allegro.ai) must be configured manually on the cloud platform of choice. For Polyaxon and Kubeflow, this means setting up and managing a Kubernetes cluster. Conveniently, once this cluster is up and running, these tools are able to automatically allocate compute resources for model training. In the case of MLFlow and allegro.ai, a central server must be set up to receive, manage, and serve experiment metadata and performance metrics. These platforms do not have the ability to automatically spin up compute resources for model training. Instead, users run training code, instrumented with platform-specific libraries, on compute which they allocate and manage themselves. In both cases, the user interface (UI) of the platform, where model performance and metadata is tracked, is only available if the Kubernetes cluster or centralized server is actively running. Here we see an advantage for AI Platform and Azure Machine Learning, as these services handle allocation of compute resources and provide an always available user interface for managing jobs, models, and artifacts.

Finally, there is diversity across model organization and evaluation features. In terms of organization, only MLFlow and Azure Machine Learning offer a true model registry where models and are named, versioned, and paired with training artifacts. Other platforms instead organize models and artifacts on a per-experiment basis; each trained model is unique, and there is no concept of "versioning" a particular one. When it comes to evaluation, the user interface plays the most important role. Kubeflow and AI Platform offer minimal functionality, focusing primarily on tracking individual training and optimization jobs rather than the trained models and metrics these jobs output. MLFlow, Polyaxon, allegro.ai, and Azure Machine Learning offer a more fully featured dashboard for analyzing model performance and comparing models against one another.

There are tradeoffs no matter which platform is utilized. For our purposes, the top priorities are cost, flexibility, and simplicity. It is desirable to avoid managing and paying for a separate server or Kubernetes cluster in the way MLFlow, Polyaxon, Kubeflow, and allegro.ai require to track experiments and compare models. Likewise, the ability of the pipeline to manage compute allocation is critical for ease of use. We also desire a pipeline that is agnostic to cloud provider for maximum utility across organizations. And, as previously mentioned, open-source is a huge plus; utilizing closed source platforms like Azure Machine Learning or AI Platform exposes the system to feature deprecation or paywalling beyond our control. Equally important, synthetic image generation is computationally expensive and benefits greatly from advanced GPUs. A flexible pipeline architecture that fully integrates cloud-based synthetic image generation is desirable. Otherwise, a separate mechanism for cloud-based image generation must be developed and maintained, including a system for compute allocation. These considerations led to the conclusion that a custom solution, leveraging existing services where possible, was the best choice for our needs.


%% file: synthetic.tex
As mentioned in Section \ref{sec:introduction}, a large quantity of diverse and representative training data is a key requirement for deep learning; however, the space environment poses unique challenges to collecting a large number of labeled training images. There is simply a lack of real images of spacecraft on-orbit, and none at all for spacecraft that have yet to launch. As such, the application of deep learning techniques to on-orbit visual navigation will necessitate the usage of synthetic data for the foreseeable future.

However, generating synthetic images in the space environment presents unique challenges of its own. Space imagery is characterized by a wide variety of lighting conditions, spacecraft poses, and dynamic Earth backgrounds. It often includes aberrations such as blur, overexposure, or lens flares, particularly when using inexpensive monocular cameras. Training robust deep learning models on purely synthetic images is no easy task, and imitating these real conditions as closely as possible is paramount. This means that a synthetic image generation framework must include flexible and easy configuration of many scene parameters. Furthermore, it must be powerful enough to emulate a wide variety of effects and aberrations. Photorealism is also a primary concern.

Our synthetic image generation framework is built on top of the 3D graphics toolset Blender. Blender affords numerous advantages that address many of the challenges stated above. It comes pre-packaged with Cycles: a physically based, ray-tracing, production quality rendering engine. In comparison with real-time rendering techniques such as Unreal Engine 4 or OpenGL-based renderers, Cycles better mimics real material and lighting characteristics. Together, Blender and Cycles are capable of modeling an innumerable variety of complex visual effects. Crucially, Blender includes a powerful Python API that enables easy control over every aspect of image generation. As an added bonus, every component of Blender is completely open-source and free to use. 

\subsection{Starfish}
\label{subsec:starfish}
The foundation of our synthetic image generation framework is Starfish, a custom-built, open-source Python library that integrates with Blender's Python API and facilitates automatic image generation. At its core, Starfish abstracts away the complexities of object and camera positioning, parameterizing an image of a single target spacecraft with six intuitive variables:

\begin{itemize}
    \item \textbf{Position:} The absolute 3D position of the target in the global coordinate system. This is only important if there are other objects in the scene that the target must be positioned relative to.
    \item \textbf{Distance:} The distance between the target and the camera, in a real unit such as meters.
    \item \textbf{Offset:} The 2D offset of the target in the camera frame.
    \item \textbf{Target orientation:} The orientation (attitude) of the target with respect to the camera's reference frame.
    \item \textbf{Background orientation:} The orientation of the camera with respect to the global reference frame; in other words, the orientation of the scene ``background'' in the camera frame.
    \item \textbf{Lighting orientation:} The direction that the sun's light comes from, with respect to the camera's reference frame.
\end{itemize}

\begin{figure}[htpb]
\centering
\includegraphics[width=\linewidth]{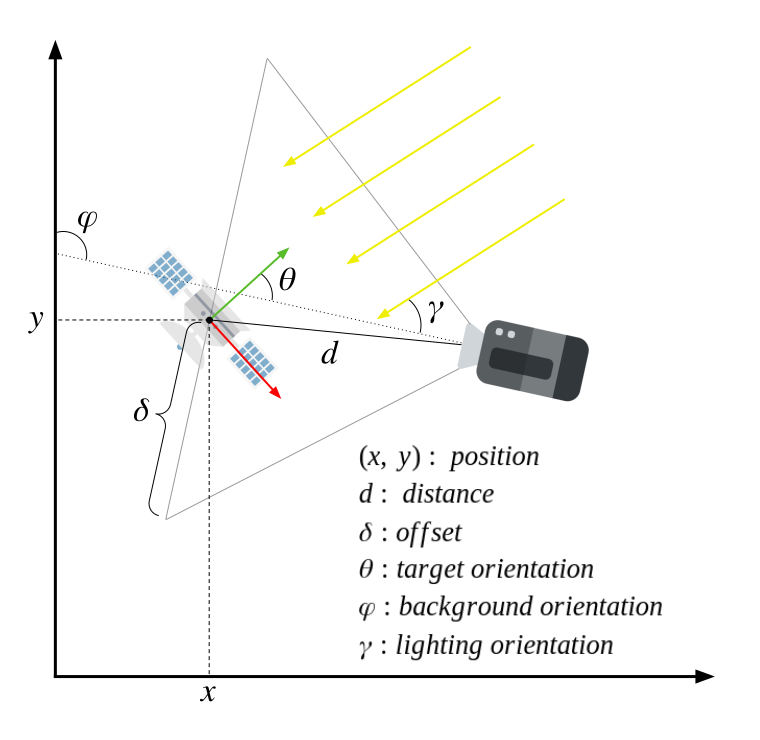}
\caption{A two-dimensional representation of the configurable image parameters in Starfish}
\label{fig:starfish}
\end{figure}

\begin{figure}[htpb]
\centering
\begin{subfigure}[t]{1.5in}
  \includegraphics[width=\linewidth]{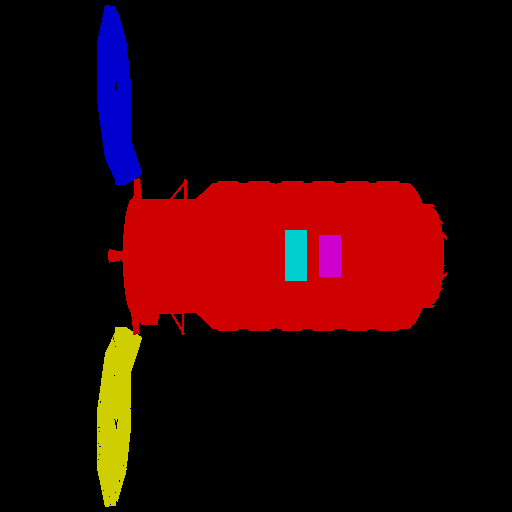}
\end{subfigure}\hfil 
\begin{subfigure}[t]{1.5in}
  \includegraphics[width=\linewidth]{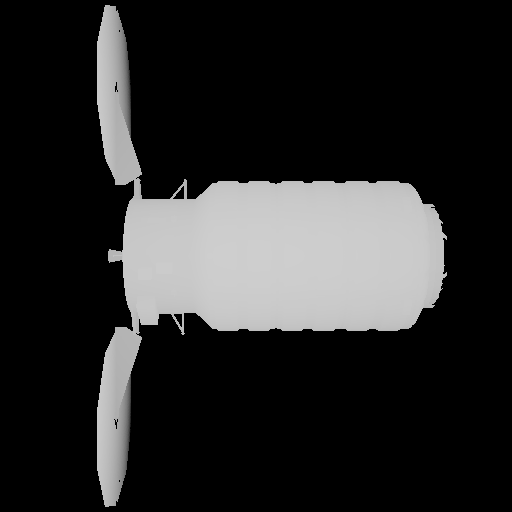}
\end{subfigure}\hfil 
\caption{Segmentation mask (left) and depth map (right) image labels}
\label{fig:mask_depth}
\end{figure}

Figure~\ref{fig:starfish} provides a simplified 2D illustration of these six parameters. Given a suitable 3D model, Starfish is able to render an image with any combination of these six parameters and output corresponding labels such as the target's relative position and attitude. With the right Blender setup, Starfish can produce additional labels such as bounding boxes, keypoint locations, segmentation masks, and depth maps as shown in Figure~\ref{fig:mask_depth}.

Building upon this single-image parameterization, Starfish provides additional functionality for creating sets of images. For training data, these sets can be randomized over ranges of the parameter space to produce a uniform distribution over desired conditions. These sets can also be constructed by smoothly interpolating between points in parameter space, which is useful for models that require time-series data containing continuous movements of the target or camera. With fine enough resolution, these images can be combined into a high quality animation of any desired movement profile. In this way it is possible to simulate a real mission profile to evaluate model performance on and identify failure cases. These animations also make great marketing material.

\subsection{Effects and Augmentations}
\begin{figure}[htpb]
    \centering 
    \begin{subfigure}[t]{1.5in}
        \includegraphics[width=\linewidth]{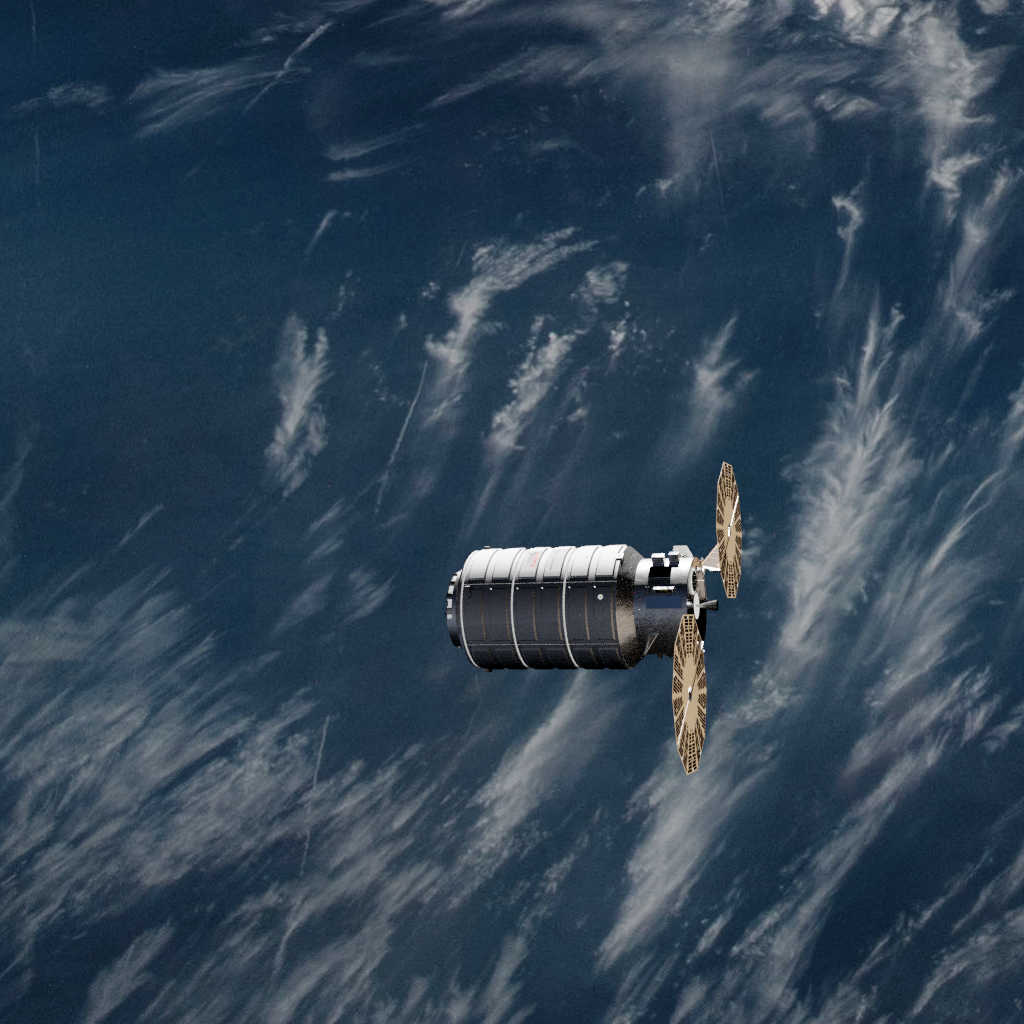}
        \caption{\normalfont{Cygnus with no augmentations in front of a real Earth background}}
    \end{subfigure}\hfil 
    \begin{subfigure}[t]{1.5in}
        \includegraphics[width=\linewidth]{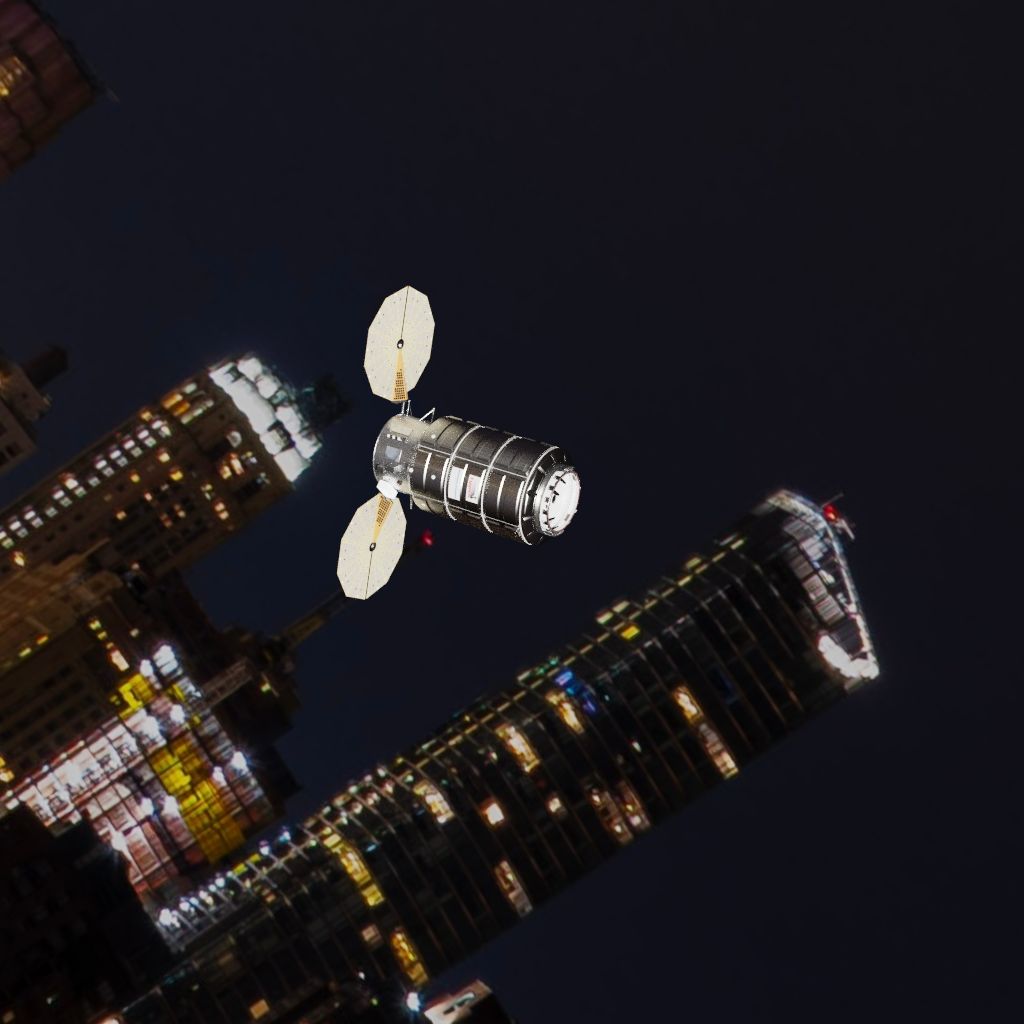}
        \caption{\normalfont{Cygnus in front of a randomized background}}
    \end{subfigure}\hfil 

    \medskip
    \begin{subfigure}[t]{1.5in}
        \includegraphics[width=\linewidth]{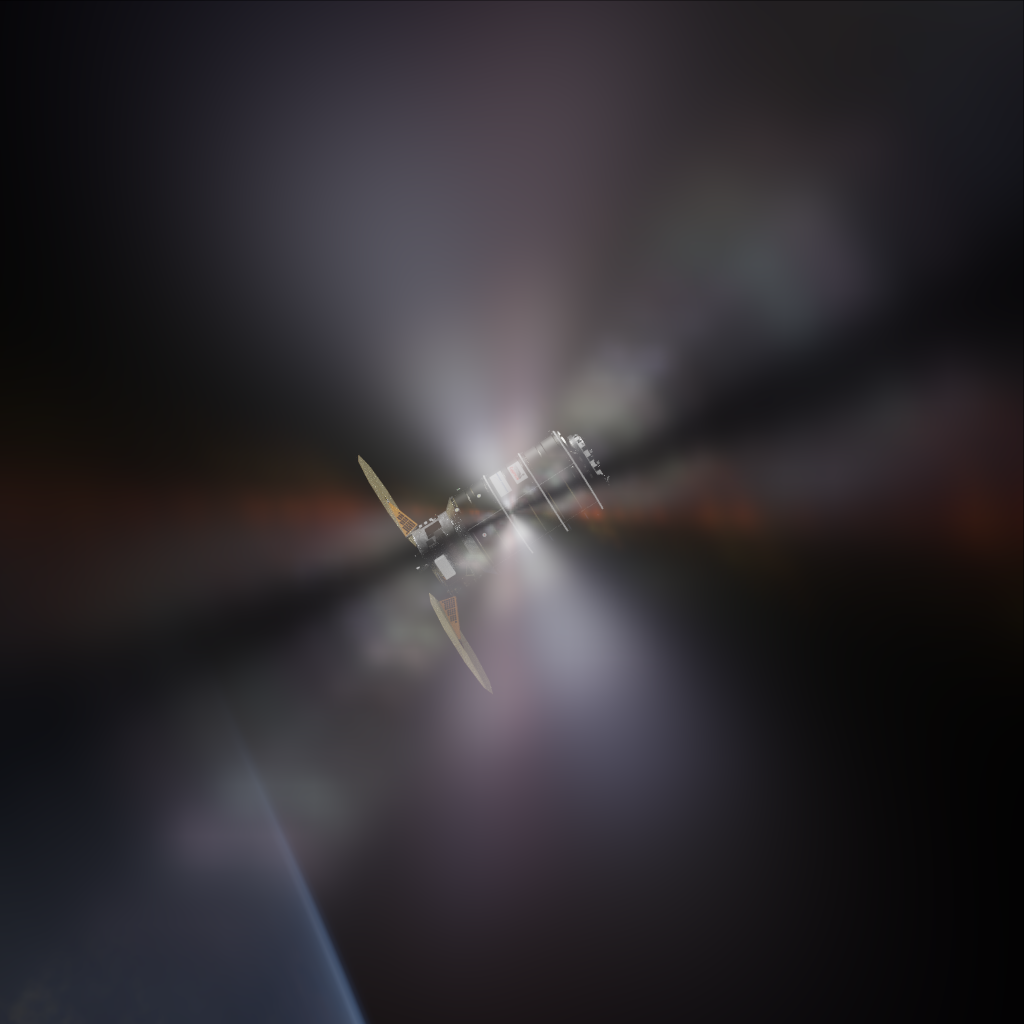}
        \caption{\normalfont{Cygnus with a lens flare effect}}
    \end{subfigure}\hfil 
    \begin{subfigure}[t]{1.5in}
        \includegraphics[width=\linewidth]{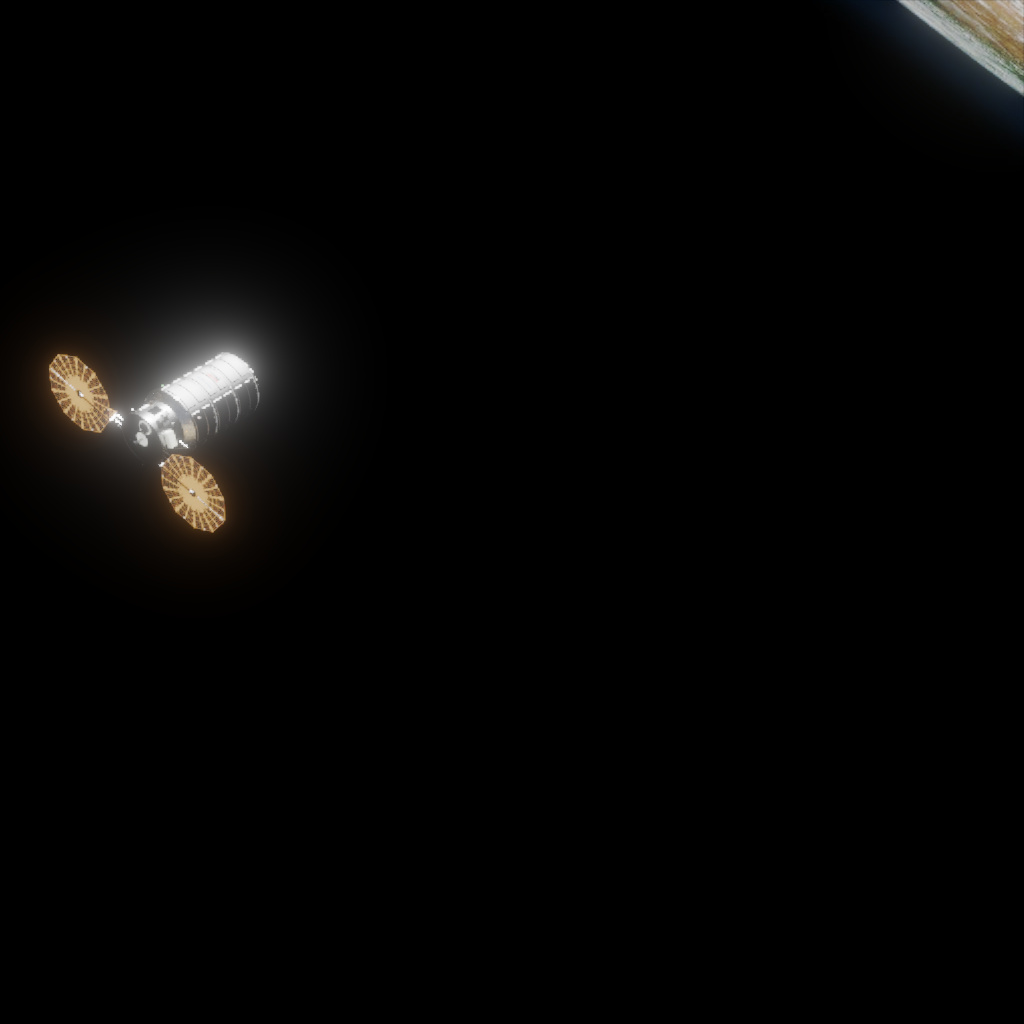}
        \caption{\normalfont{Cygnus with a ``fog glow'' effect}}
    \end{subfigure}\hfil 

    \medskip
    \begin{subfigure}[t]{1.5in}
        \includegraphics[width=\linewidth]{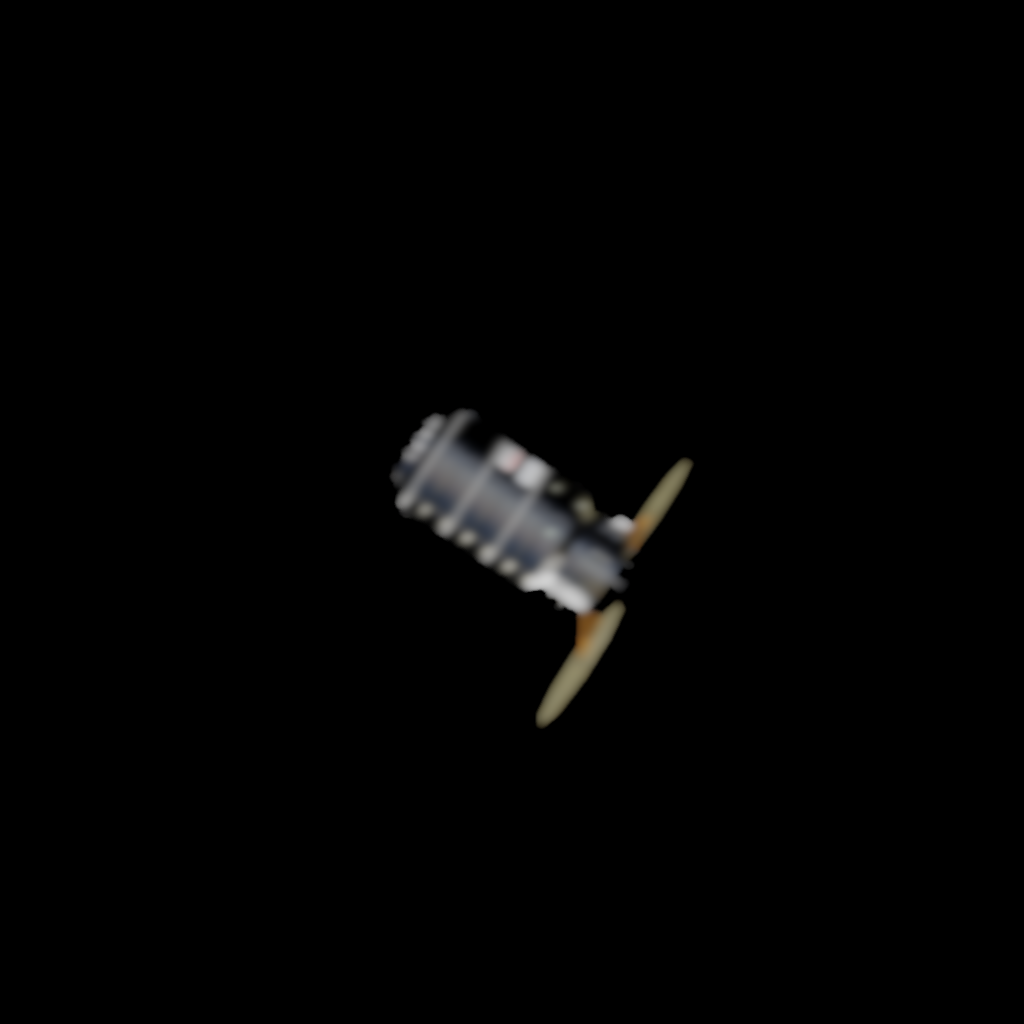}
        \caption{\normalfont{Cygnus with a blur effect}}
    \end{subfigure}
    \begin{subfigure}[t]{1.5in}
        \includegraphics[width=\linewidth]{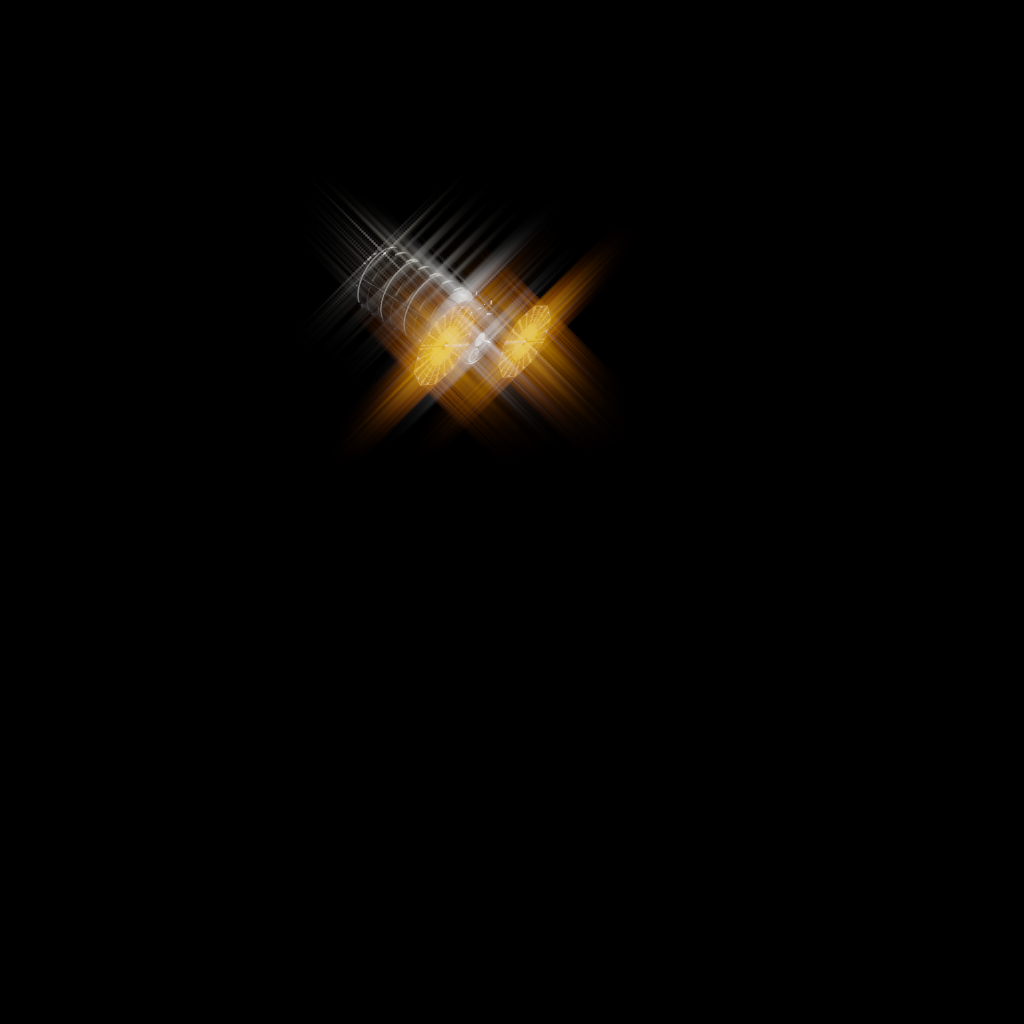}
        \caption{\normalfont{Cygnus with a ``simple star'' effect}}
    \end{subfigure}
    \caption{Example synthetic images with and without augmentations.}
    \label{fig:augmentations}
\end{figure}

As mentioned above, Blender is capable of producing a wide variety of high-quality visual effects. On top of Starfish, we use Blender's Compositing tool to replicate several observed in-space aberrations, particularly those that led to detection failures during the Seeker-1 mission. These effects primarily include lens flares, bloom, and blur. The presence and strength of these augmentations is easily configured using Blender's Python API, integrating seamlessly with automatic image generation. Figure~\ref{fig:augmentations} provides some examples of typical augmentations.

We also use the Compositing tool to replace the backgrounds of some images from the default pure-black space background. Two types of backgrounds are used. The first are real satellite images of Earth, which are abundant and thus easily used to improve photorealism in comparison to a synthetic Earth background. The second are completely randomized non-space images from various datasets. These backgrounds are a form of domain randomization, where certain irrelevant elements of training data are randomized in a way that does not affect the ground-truth target labels. Domain randomization has been shown to help reduce model overfitting and improve simulation-to-real transfer \cite{DomainRandomization, Tremblay2018}.

\subsection{Pipeline Integration}
\label{subsec:pipeline_integration_synth}
Using Starfish and Blender, a wide variety of imagesets --- our term for sets of related labeled images --- of any size can be crafted using a few dozen lines of Python. As mentioned above, Starfish can output various pieces of data with each image. Labels like segmentation masks and depth maps are exported as a separate file for easy viewing, while all other data is exported in a single Javascript Object Notation (JSON) metadata file. Starfish allows users to optionally set additional metadata fields like generation sequence name, timestamp, and most importantly, identifying tags. Each image can have an arbitrary number of tags that characterize it in some way. For example, an image of Cygnus over a cloudy earth background may have the tags "Cygnus" and "cloudy earth." These enable rich downstream filtering for dataset creation purposes. After an imageset is finished generating, we create an imageset-wide metadata file with key identifying information like name, author, and git commit, and then automatically upload the set to cloud storage for ingestion by the rest of our pipeline. Uploaded imagesets are organized by name in the cloud.

As we discussed in Section \ref{sec:existing_tools}, synthetic image generation is computationally expensive. We leverage Amazon EC2 to run generation sequences on powerful compute instances that drive down runtimes. On an instance equipped with one NVIDIA Tesla M60 GPU (g3.4xl), a single $1024 \times 1024$ image takes approximately 8 seconds to generate with our Blender configuration. A set of 10,000 images can thus be generated and available in cloud storage in less than a day using relatively inexpensive cloud resources: the cost to generate such a dataset would be just over \$25 at current prices \cite{ec2_pricing}. 

It is important to note that every step after actual image generation is targeted at integration with the rest of our pipeline and is tuned to our use cases. While we have found this particular scheme valuable, users are free to adopt their own patterns and leverage metadata, cloud resources, and other features as they see fit. As we will describe in Section \ref{sec:pipeline}, organization of labeled images within each imageset in the cloud is arbitrary when used downstream in the pipeline, and a lack of descriptive tags or imageset names simply results in a less feature-rich dataset creation experience.

%% file: pipeline.tex
\begin{figure*}[htpb]
    \centering
    \includegraphics[width=7in, keepaspectratio]{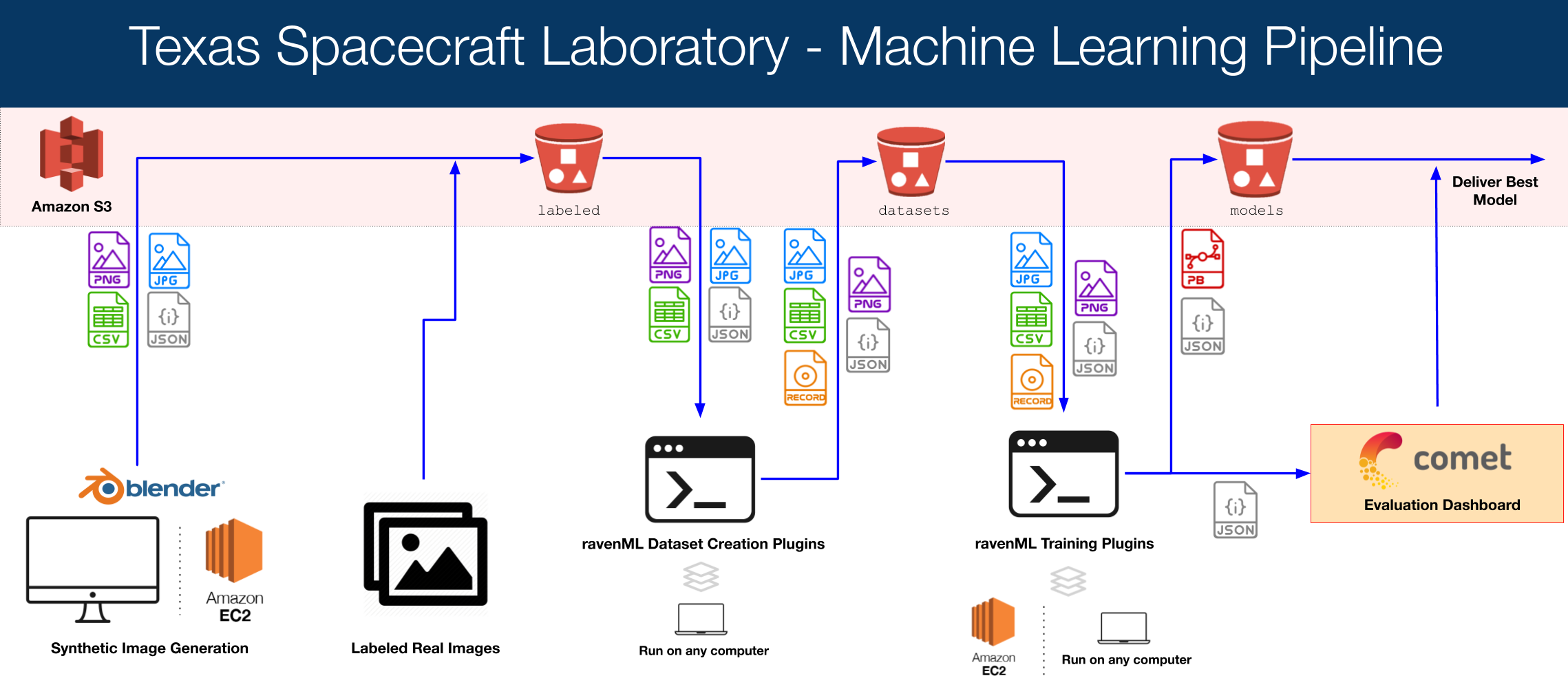}
    \caption{Full pipeline flowchart.}
    \label{fig:pipeline_diagram}
\end{figure*}

The model development pipeline itself consists of four high level tasks: synthetic image generation, dataset creation, model training, and evaluation. Each stage flows data downstream through shared cloud storage and is managed by a specific software tool. The full pipeline is shown in Figure \ref{fig:pipeline_diagram}. 

Two custom open-source tools, Starfish and ravenML, combine to manage the first three stages in the pipeline. Both are Python packages installable with pip, the standard package manager for Python. Starfish handles image generation and is described previously in Section \ref{subsec:starfish}. Dataset creation and model training are handled by ravenML. Finally, an enterprise solution for model evaluation, cometML \cite{cometML} --- available free for academics --- provides an evaluation dashboard for metric visualization and logging.

Crucially, the entire pipeline can be leveraged from any internet-connected machine, and there is no limit to the number of concurrent image generation or model training jobs. Use of cloud storage means all developers have instant access to the latest available images, datasets, trained models, and performance metrics produced by these runs. This data is logically organized by each tool that generates it, and is always paired with associated metadata such as author, git commit hash, and other vital information for system comprehension and reproducibility.

\subsection{Cloud Integration}
\label{subsec:cloud}
Since many aerospace activities are subject to International Traffic in Arms Regulations (ITAR), it is critical that any third-party services present in the pipeline be ITAR-compliant. Amazon Web Services (AWS) \cite{aws} was selected for cloud services due to its GovCloud \cite{aws_gov} platform, which offers the vast majority of AWS services from US-based, ITAR-compliant data centers.  We utilize two AWS services: Simple Storage Service (S3) and Elastic Compute Cloud (EC2).

S3 use is shown in Figure \ref{fig:pipeline_diagram} as the red bar across the top. The top level of storage in S3 is referred to as a bucket. Three buckets are used, one for each stage of data in model development: raw labeled images, curated datasets, and trained models (including training artifacts). Each bucket is logically organized and managed by the tool that uploads data to it, and downstream tools understand and use this structure to grab required data. All of this occurs automatically behind-the-scenes to provide users the data they need without any management effort on their part. They simply request data by some identifying name, and the tool knows where to find it. If the data needs to be accessed manually at any time, rather than by a tool in the pipeline, it is simple to do so thanks to this automatic organization.

EC2 is used to execute the long-running, GPU intensive steps of synthetic image generation and model training. Necessary compute resources are spun up by a simple command line tool, nicknamed bran, which users run on their local machine. bran starts and configures a compute instance for the desired task by installing ravenML or Starfish and opens a terminal connection to the running instance. Users then trigger the desired operation from this terminal. At this point, no further interaction is necessary --- the job runs until completion, at which time the tool (either Starfish or ravenML) automatically shuts down the compute instance using the AWS Python Standard Development Kit (SDK). This ensures users only pay for the precise amount of compute they use.

There is one huge benefit to this setup: consistency between the local and cloud experiences. In both cases, ravenML and Starfish are simply installed into Python virtual environments and used from the terminal. There is no difference in how a user interacts with the tool, and local setup is as simple as installing a few Python packages. This duality makes testing and debugging a breeze. For cases where locally available hardware meets project needs, organizations can simply use the presented pipeline entirely on this hardware and still gain all pipeline benefits like data organization and metadata generation. Or, as we have done ourselves, a hybrid approach can be taken where locally available hardware is used when possible and EC2 serves as a backup whenever needs increase. Such flexibility can provide appreciable cost savings, and the utility of simply installing, using, and developing project code anywhere cannot be overstated.

\subsection{ravenML}
\label{subsec:ravenML}
\begin{figure*}[htpb]
    \centering
    \includegraphics[width=7in, keepaspectratio]{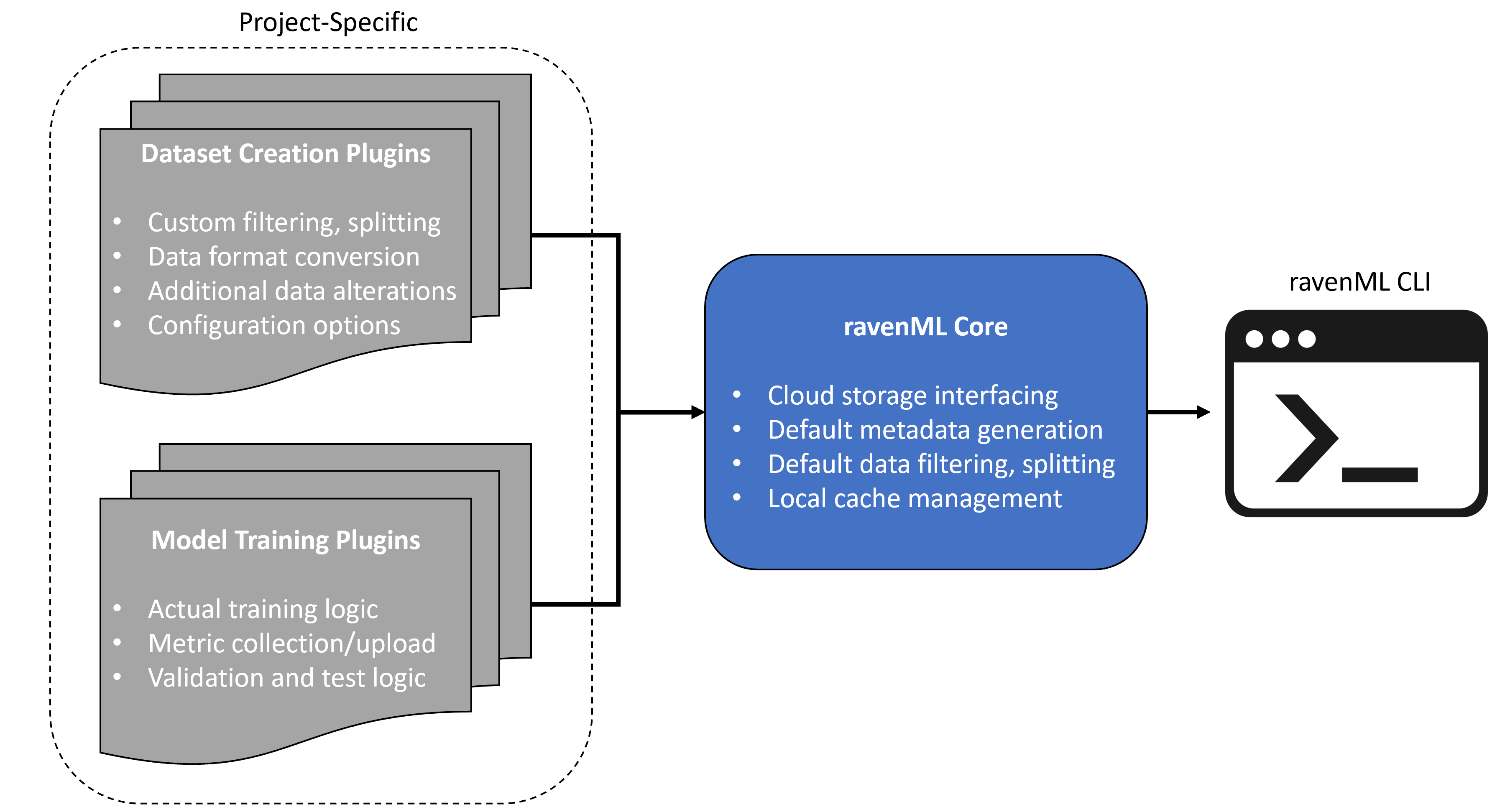}
    \caption{ravenML architecture.}
    \label{fig:ravenML}
\end{figure*}

A custom software tool written in Python, ravenML, is used for both dataset creation and model training. This functionality is exposed through an extensible command line interface (CLI). For each project developers write two plugins, one for dataset creation and another for model training. Plugins are entirely independent and can implement these operations using any desired libraries or techniques. In turn, ravenML provides core functionality common to all projects such as metadata generation, cloud storage interactions, and cache management. Plugins are required to be pip-installable Python packages; thus, their structure is similar regardless of project, providing a template to help keep code organized and readable. Figure \ref{fig:ravenML} describes the ravenML architecture.

The plugin system enforces a base level of consistency across projects, resulting in a similar user experience regardless of plugin author or model type. It is always true that:

\begin{itemize}
    \item[(1)] the plugin is a pip-installable Python package \\
    \item[(2)] dataset creation and model training is kicked off by calling a single plugin command via the ravenML CLI \\
    \item[(3)] dataset creation and model training configuration is passed in via YAML files
\end{itemize}

YAML files are used for configuration, rather than command line arguments, as deep learning often has dozens of configuration options. ravenML defines a base set of fields required for all dataset creation and model training operations. Plugins receive necessary input through independent configuration fields within the YAML. Plugin configuration can thus be arbitrarily complex and include additional metadata fields if desired. ravenML handles loading and parsing these configuration files before passing them internally to the plugins. Example YAML files are shown in subsequent sections.

Dependency tracking is paramount for efficient model development and reproducibility. Since all plugins are pip-installable Python packages, developers are forced to maintain a list of dependencies alongside their plugin. This simplifies environment setup, whether locally or on the cloud, to a few simple pip commands. No matter the project, team members can always expect to utilize these simple commands to set up and use dataset creation and model training code.

ravenML also provides optional utility modules for operations common across plugins to reduce code duplication within teams and simplify codebase maintenance. The most prominent of these is a user interfacing module, also used internally by ravenML itself. Plugins that use this module can expose an interface that mirrors the core ravenML CLI.

\subsection{Dataset Creation with ravenML}
\label{subsec: dataset_creation}
\begin{figure*}[htpb]
    \centering
    \includegraphics[width=7in, keepaspectratio]{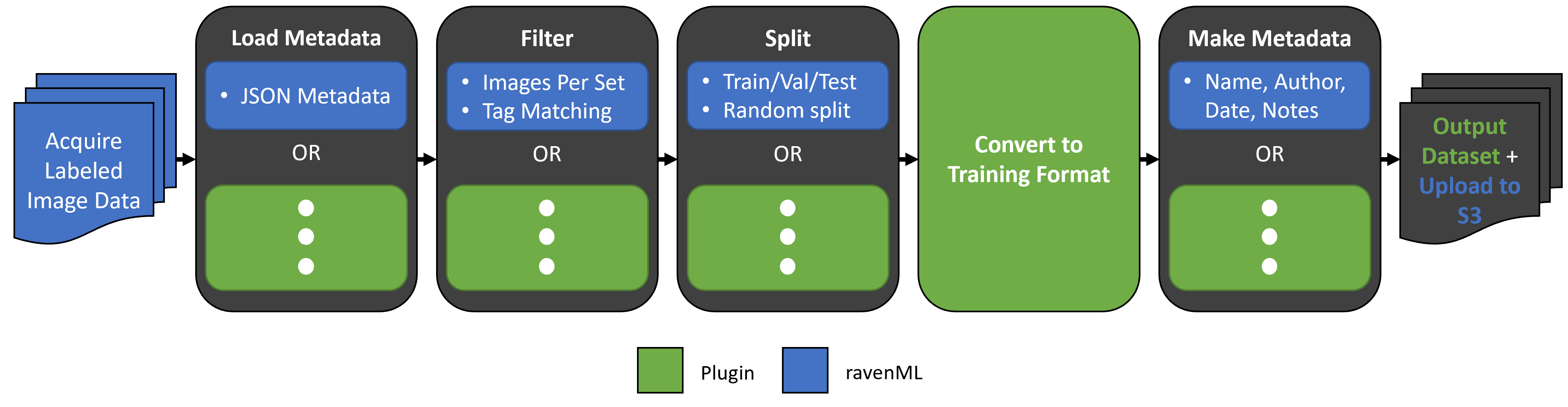}
    \caption{ravenML dataset creation stages.}
    \label{fig:rml_dataset_creation}
\end{figure*}

Dataset creation takes raw, labeled images and prepares them for model training. ravenML defines seven steps for dataset creation:

\begin{itemize}
    \item[(1)] Define labeled image source (either S3 or local) \\
    \item[(2)] Load per-image metadata \\
    \item[(3)] Filter images in the source by their metadata \\
    \item[(4)] Split remaining data into train, validation, and test sets \\
    \item[(5)] Convert images and labels into training data format \\
    \item[(6)] Collect dataset metadata \\
    \item[(7)] Output dataset and metadata, and optionally upload to S3
\end{itemize}

ravenML offers a convenient way to handle all of these tasks at once through the CLI. It provides default implementations for all but (5). Thus, only the final step of data conversion must be implemented by project-specific plugins. Plugins can optionally override any of (2-7) wherever custom logic is required. For example, a plugin may prefer a different dataset directory structure, and so would override the output step in (7). Figure \ref{fig:rml_dataset_creation} summarizes the dataset creation process.

When sourcing from S3, ravenML assumes that the S3 bucket is organized into individual imagesets (defined in Section \ref{subsec:pipeline_integration_synth}). Users then select the desired imagesets to include in their dataset. This is the only requirement on labeled image organization; the contents of each imageset can take on any desired structure, so long as the implementation for (2) can interpret it. The default implementation provided by ravenML is designed for Starfish integration and so interprets imagesets according to the Starfish output format.

Choice of imageset is the first step in a rich sequence of data filtering. As the composition of deep learning datasets plays a critical role in model performance, filtering is a key operation. The default ravenML implementation provides two types of filtering: by size and by tag. Size filtering allows selection of a precise number of images from each source imageset. Tag filtering allows users to select one or more tags and define either an AND or OR condition with these tags. For example, selecting the tags "Cygnus" and "cloudy" with an AND condition would select only images with both tags, while using an OR condition would select all images with at least one of the tags. Multiple tag filters can be used in one filtering operation. Each is independent and defines a unique subset of the entire set of available images. When the user is done making subsets with the tag filter, all subsets are merged, and users are given the option to use only a specific number of these images for the final dataset. Size and tag filtering can be used in sequence; in the default implementation, size filtering is executed first if used.


The default filtering implementation makes piecing together datasets from a large number of pre-existing imagesets simple and intuitive. We have found that over time, projects accumulate a large amount of labeled data, especially when using synthetic imagery that can be constantly generated. ravenML keeps all this labeled data organized and easy to use, allowing developers to focus solely on dataset composition. While the default implementation has been sufficient for our needs, dataset plugins are free to override these features with arbitrarily complex functionality.

An example YAML configuration file for dataset creation is shown in Listing \ref{listing: yaml_dataset_input}. It creates a dataset named \lstinline{tf_object_detection_dataset} using images from the \lstinline{good_lighting}, \lstinline{bad_lighting}, and \lstinline{cloudy_backgrounds} imagesets with a 20\% train-test split.

\begin{lstlisting}[float=htpb, caption={Example dataset creation YAML configuration.},label={listing: yaml_dataset_input}]
dataset_name: tf_object_detection_dataset
local: False
imageset:
  - good_lighting
  - bad_lighting
  - cloudy_backgrounds
overwrite_local: true
kfolds: 5
test_percent: .2
upload: True
delete_local: True
metadata:
    created_by: The Author
    comments: Best dataset ever
plugin:
    verbose: true
    filter: false
\end{lstlisting}

\subsection{Model Training with ravenML}
\label{subsec: model_training}
Model training is handled by separate project-specific training plugins. The only requirement of these plugins is that they expose a single command which invokes model training and outputs a trained model in the format ravenML core expects for upload to S3. There are no limitations to the code that can be executed by this command as long as it satisfies these requirements; any deep learning framework can be used. Plugins can also define additional commands for specific types of model evaluation or visualization.

ravenML uses a flexible structure to interpret model outputs. When a plugin finishes training a model, it provides two paths to ravenML core: one to the trained model file itself, and another to a directory of extra files associated with the model like training checkpoints, label maps, and anything else the plugin developer deems important. ravenML generates a unique ID for the model and, on upload, renames the trained model file with this ID as a suffix to the user configured model name. Metadata files also use this ID as a suffix. These two files are placed into the models bucket under a common prefix containing all trained models and metadata. A separate folder, named with the unique ID, is then created under a separate prefix within the bucket to store all extra files given by the plugin. In this way, no matter where a model file ends up --- whether integrated deep into a system, or loose on a developer's computer --- it can be tied to its training metadata and any important associated artifacts.

As stated earlier, we use cometML for model evaluation and tracking. This means plugins we have created are instrumented with cometML's Python API to stream logs, metrics, and other assets to the cometML server during training. However, plugins are free to use any library or service they wish for this purpose, or none at all. ravenML guarantees that the trained model and artifacts will be safely stored on S3; beyond that, it is up to teams how they wish to visualize and evaluate model performance. Services like neptune.ai \cite{neptune.ai}, the aforementioned allegro.ai, and others can easily be swapped in for cometML. Organizations can even run these services in parallel, using the one best suited for each particular project.

An example YAML configuration file for model training is shown in Listing \ref{listing: yaml_train_input}.  It trains a MobileNetV2 Single Shot Detector model on the \lstinline{aeroconf} dataset over 1000 training steps. This model and associated artifacts will automatically be uploaded to the models S3 bucket at the conclusion of training.

\begin{lstlisting}[float=htpb, caption={Example model training YAML configuration.},label={listing: yaml_train_input}]
dataset: aeroconf
overwrite_local: True
ec2_policy: stop
metadata:
    created_by: The Author
    comments: Best model ever
plugin:
    verbose: true
    comet: true
    model: ssd_mobilenet_v2_coco
    optimizer: RMSProp
    hyperparameters:
        train_steps: 1000
\end{lstlisting}

\subsection{Model Evaluation and Comparison Using cometML}
\label{subsec:evaluation}
We chose to leverage cometML for model evaluation and comparison because it is free to academics and is hosted externally, requiring no server setup or maintenance. While S3 serves as the source of truth for trained models, checkpoints, and metadata, cometML allows exploration of these models in a user-friendly way both during and after training. As touched on above, the cometML Python API is used to stream data to the cometML dashboard. These data include training information like hyperparameters, accuracy and loss, and assets like images or configuration files. Additionally, the API can be used to log all operating system and Python packages installed at the time of training. It also automatically tracks system metrics like GPU and memory utilization to give a comprehensive view of the training process. Each single training job is a unique object in the cometML dashboard and is assigned to a specific project.

\begin{figure}[htpb]
    \centering
    \includegraphics[width=3in, keepaspectratio]{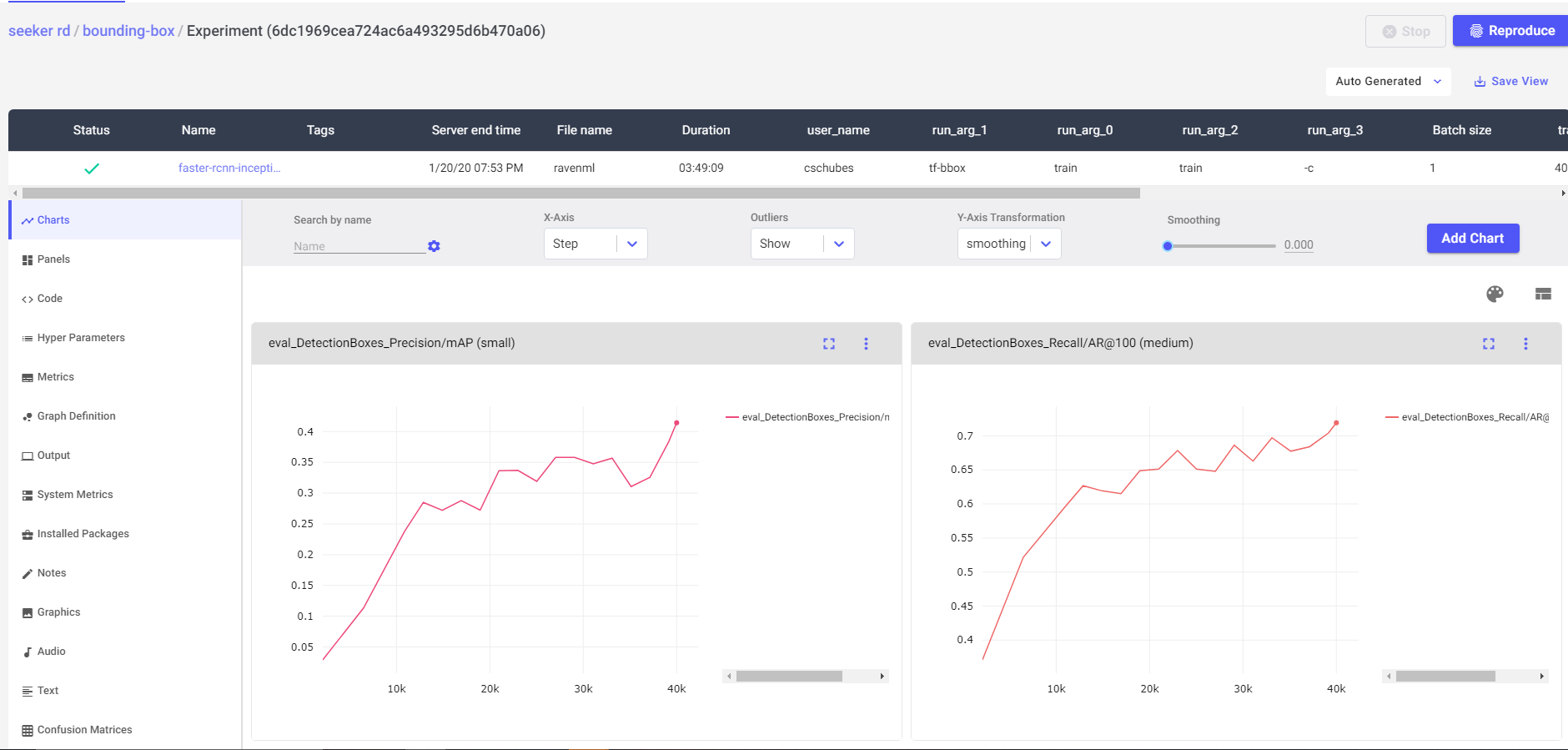}
    \caption{cometML training metrics.}
    \label{fig:cometML}
\end{figure}

\begin{figure}[htpb]
    \centering
    \includegraphics[width=3in, keepaspectratio]{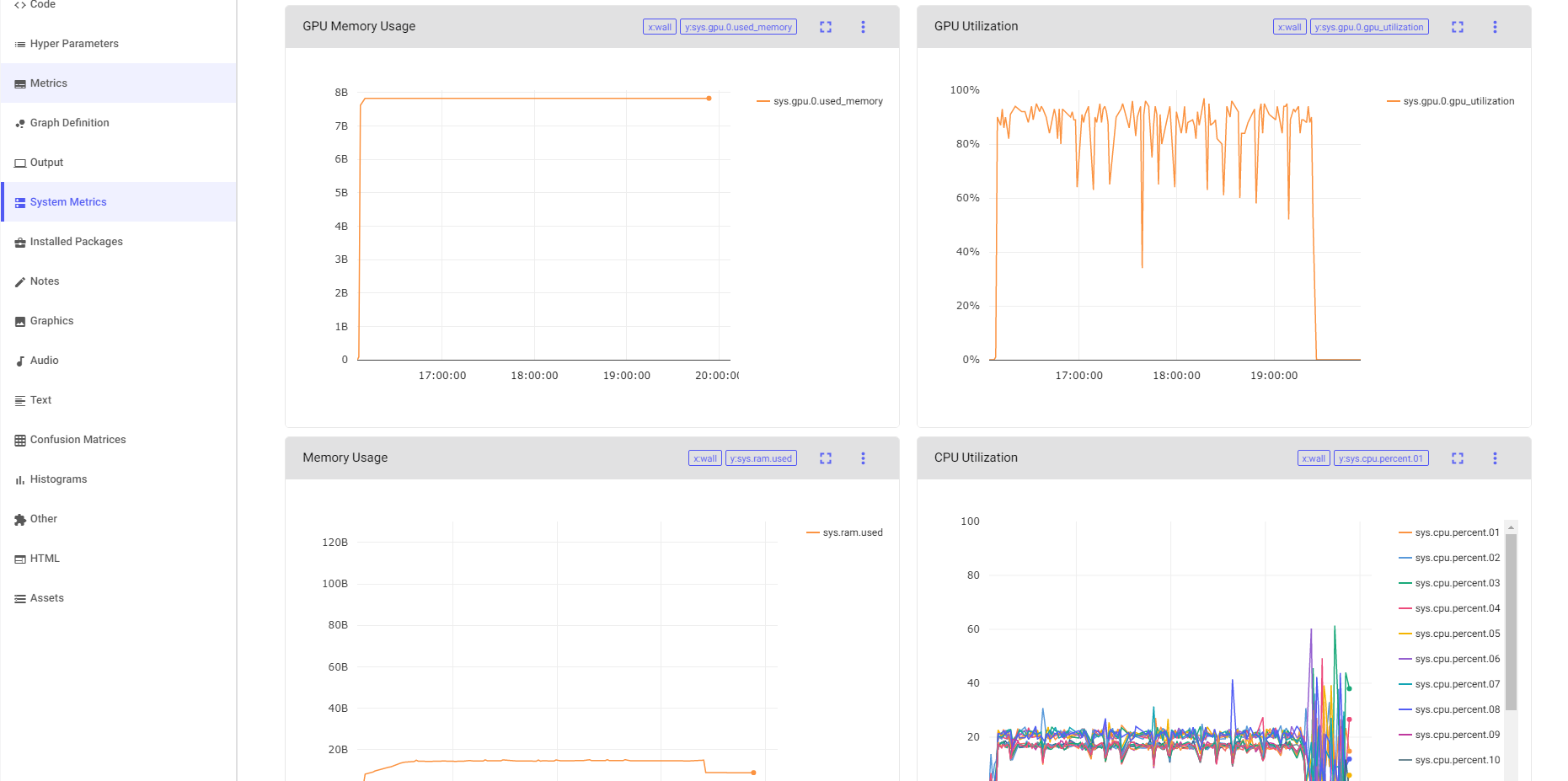}
    \caption{cometML system metrics.}
    \label{fig:cometML_memory}
\end{figure}

Figure \ref{fig:cometML} shows the dashboard view of an experiment's visualized training metrics. Figure \ref{fig:cometML_memory} shows the dashboard view of an experiment's system metrics.

%% file: case_examples.tex
In the following sections we present examples of pipeline use that demonstrate its benefits for model development, system performance, and flexibility across different mission profiles. 


\subsection{NASA Gateway}
\label{subsec:gateway}
Object detection research in the space domain is complicated by the fact that most spacecraft are unique, and different machine learning models must be developed to detect different spacecraft. One key strength of the pipeline is the ease with which researchers can change their target spacecraft in an efficient and organized manner. We present a scenario in which the agility provided by this pipeline allowed for a quick research turnaround. 


The authors were tasked with developing a relative bearing estimation system with the same capabilities as the Seeker Vision system for application to the proposed Lunar Gateway space station. The motivation for this research was to explore the possibility of using visual navigation as part of a suite of navigation options to guide the Orion spacecraft to the Lunar Gateway station. The time constraint on the development of this model was extremely tight, a few weeks compared to the months of development taken for the original model built for the Cygnus spacecraft. This compressed timeline required the team to rely heavily on the pipeline. Additionally, the hardware used for the project was an Intel NUC, a far more powerful platform than the Intel Joule used for the Seeker Vision project. 


As a result of using the pipeline, the only substantive change that had to be made was importing the new 3D Gateway Blender model provided to us by NASA into a new Blender file in order to generate synthetic imagery. Using Starfish and Blender, the team generated around twenty thousand synthetic images of the Lunar Gateway in a variety of poses and lighting conditions. These images, and their related metadata, were organized into two training datasets with the entire creation process taking less than 24 hours.

As the Intel NUC provided for this research is far more computationally powerful than the hardware used in the Seeker Vision project, the pool of possible object detection architectures that would satisfy speed requirements was much larger than in the case of Seeker Vision. The ravenML bounding box detection plugin allowed the team to efficiently train and evaluate over two dozen deep learning models using a variety of model architectures in a period of only two weeks. 

After this period of rapid development and iteration, the team produced a model based on the Faster-RCNN architecture capable of localizing the Lunar Gateway space station in synthetic monocular imagery taken of the station. The speed with which such a model, and the infrastructure necessary to train it, was created was due to the streamlined development process facilitated by the pipeline. Starfish allowed the team to automate the generation of a large amount of synthetic imagery in a variety of poses, lighting angles, and backgrounds with the user responsible only for providing an initial Blender model of the target spacecraft. ravenML and bran were then used to quickly kick off model trainings in the cloud, with ravenML allowing the user to easily change parameters and options for each training. These tools, along with the cloud resources that link them together, allowed the group to easily iterate on their trained models and develop a model capable of effectively performing the task of detecting the Lunar Gateway from monocular imagery in under a month.

\subsection{Bounding Box Plugin}
\label{subsec:odplugin}
One common problem in vision-based on-orbit proximity operations is the need to identify and localize an object in an image, a task known as object detection. A common method of performing this localization is known as bounding box object detection, wherein the coordinates of a rectangular box tightly surrounding the object of interest are returned as an output. In recent years, deep learning based object detection models have proven to perform well at this task. A powerful existing tool for creating deep learning based bounding box detection models is the Tensorflow Object Detection API \cite{Huang_2017}. This framework allows users to choose from a variety of state of the art model architectures without having to build them from scratch. However, sparse documentation and a complicated training scheme render the tool untenable for rapid, efficient, and organized model development without additional tooling. The authors thus developed a ravenML training plugin in order to allow users to train bounding box models in a simple way. The bounding box plugin fits within the ravenML framework, providing users with the same easy-to-use interface that accompany all ravenML plugins, while also preserving the power of the Tensorflow Object Detection API. It accomplishes this with two key pieces of functionality.


The Tensorflow Object Detection API executes model trainings based on configuration files, specifying all minute details about a model including the architecture and optimizer to be used, and values for each hyperparameter. Modifying these files can be time consuming and complicated; therefore, the plugin automatically generates a Tensorflow Object Detection configuration file based on the ravenML YAML configuration file structure common to all plugins. This means users already familiar with ravenML's configuration scheme do not need to learn the Tensorflow Object Detection configuration scheme to use its advanced features.

Training a model using the Tensorflow Object Detection API requires the user to create directories that store model artifacts and data generated during the training. The onus is on the user to keep track of all of this data and ensure it is organized for future research and model reproducibility. The bounding box plugin leverages the power of ravenML to keep track of all of this data automatically. All model artifacts and data are handled by ravenML and either stored locally in an organized format or uploaded to cloud storage, depending on user preference. This functionality allows the researcher to focus less on the tedium of organizing data, and more on developing more effective models. 

\subsection{Pose Estimation Research}
\label{subsec:pose_estimation}
After the completion of the Seeker Vision project, the authors began development towards the more ambitious goal of full pose (position and attitude) estimation. Full pose estimation is a key requirement for more advanced space robotics and autonomous navigation tasks, yet is a much more challenging problem than 2D localization. The increased complexity and difficulty of creating pose estimation models was in fact a primary motivator for the initial development of the pipeline.

While many deep learning-based solutions for pose estimation exist, the subject is not nearly as well-developed as object detection and there are no existing tools for model training comparable to the TensorFlow Object Detection API. As such, pose estimation research requires a much larger code base of custom-built models that needs to be rapidly iterated upon. Additionally, many state-of-the-art techniques include object detection as a subsystem, where the target spacecraft is first localized in the image and the rest of the system operates on a cropped region of interest. This means at least two neural networks need to be trained, evaluated, and fine-tuned in conjunction.

Furthermore, the difficult nature of pose estimation makes it more sensitive to finer image details such as noise, lighting, and surface texture. In comparison to object detection, all of the aforementioned concerns about the quality and variety of synthetic training data are multiplied tenfold.

It was only thanks to the existence of the presented pipeline that the pose estimation research was ultimately successful. Multiple methods were explored, including direct pose regression, instance segmentation, and keypoint regression approaches. Each approach required new dataset creation, training, and evaluation code, all of which fit easily into the ravenML plugin framework without having to repeat much common functionality. The results of experiments with varying hyperparameters, model architectures, and dataset properties were easily tracked and organized. Over the course of 6 months, more than 100 pose estimation models and more than 100 bounding box detection models were trained on various combinations of more than 50 synthetic datasets totalling over 150,000 images. The addition of realistic effects and augmentations to the synthetic data markedly improved the ability of the resulting models to generalize to real images.

\subsection{Synthetic Data Performance Comparison}
\label{subsec: model_performance}
As a demonstration of the effectiveness of the Blender-based synthetic image generation techniques, this section presents a comparison between the original object detection model from the Seeker mission and a new model trained with Blender-based synthetic data. The former was trained on a dataset of 2,644 images, a mix between real images and Unreal-generated synthetic images. The latter was trained on all the same real images with an additional 1,300 Blender-generated synthetic images, all containing the various augmentations discussed in Section~\ref{sec:image_gen}.

The localization ability of each model is measured using the standard intersection-over-union (IoU) metric between the detected and ground-truth bounding boxes. An IoU threshold of 0.75 is established, meaning a detection is considered a true positive if and only if it exceeds the confidence threshold, the target spacecraft is in the image, and the IoU between the detected and ground-truth bounding boxes is at least 0.75. Then, at a given confidence threshold, the following binary classification metrics are computed:
\begin{equation}
    accuracy = \frac{TP + TN}{\#~images}
\end{equation}
\begin{equation}
    precision = \frac{TP}{TP + FP}
\end{equation}
\begin{equation}
    recall = \frac{TP}{TP + FN}
\end{equation}
Where TP, TN, FP, and FN stand for true positives, true negatives, false positives, and false negatives, respectively. Mean IoU is also computed across all images where the detection exceeds the confidence threshold and the target spacecraft is in the image.

The two models are each evaluated on two sets of real images. The first is the original validation set from the Seeker mission, which consists of 350 images, 200 of which contain the target spacecraft. This set is used to determine a confidence threshold that maximizes accuracy, which is 30\% for the old model and 50\% for the new model. Then, each model is evaluated on the 210 inflight images captured during the Seeker mission, all of which contain the target spacecraft.

\begin{table}[h]
\renewcommand{\arraystretch}{1.3}
\caption{\bf Validation Set Performance}
\label{tab:synthetic_val_performance}
\centering
\begin{tabular}{l | l | l}
 & Old (Real) & New (Real + Blender)\\
\hline
Accuracy & 0.93 & 0.92\\
Precision & 0.89 & 0.87\\
Recall & 0.93 & 0.93\\
Mean IoU & 0.90 & 0.89\\
\end{tabular}
\end{table}

\begin{table}[h]
\renewcommand{\arraystretch}{1.3}
\caption{\bf Inflight Image Performance}
\label{tab:synthetic_postflight_performance}
\centering
\begin{tabular}{l | l | l}
 & Old (Real) & New (Real + Blender)\\
\hline
Accuracy & 0.52 & 0.75\\
Precision & 0.78 & 0.99\\
Recall & 0.52 & 0.75\\
Mean IoU & 0.81 & 0.90\\
\end{tabular}
\end{table}

The evaluation results are documented in Tables \ref{tab:synthetic_val_performance} and \ref{tab:synthetic_postflight_performance}. Both models perform well on the validation set, which is expected since it contains many of the same types of images as the training set. However, as mentioned previously, the actual inflight images contained many unexpected aberrations (e.g., Figure~\ref{fig:skr1_inflight}) that caused the original Seeker model to struggle. The addition of Blender-generated augmented synthetic data to the training set benefits the model immensely, causing it to detect the spacecraft more often and more accurately in difficult conditions.

%% file: conclusion.tex
It is clear that deep learning shows great promise for vision-based on-orbit proximity operations. However, successful application of these techniques requires high quality synthetic data for training and an efficient model development system applicable to a broad class of spacecraft and missions. The end-to-end pipeline we have developed addresses these challenges. Synthetic data generated using Starfish and Blender, and rendered with Cycles, provides the fidelity and diversity necessary for generalization to the real environment. AWS integration automates data management and provides access to the powerful computational resources necessary for training. Finally, ravenML provides a simple, plugin-based CLI for dataset creation and model training that handles boilerplate tasks and implements a default suite of utilities like data filtering. Models can then be easily evaluated using the cometML dashboard.

While there are many existing platforms to streamline deep learning model development, the current state of deep learning for on-orbit proximity operations necessitates a low-cost, flexible solution. The presented pipeline is open-source, requires no server or cluster management, and only incurs charges for the precise amount of cloud resources utilized. It can be used locally or in the cloud, and switching between these two is simple thanks to the consistency between the two experiences.

The Seeker Vision object detection model was compared against a new objection detection model of the same architecture that was trained on an additional set of images generated with the presented Blender-based system. Model performance was compared on the 210 images captured during the Seeker 1 flight, and results clearly demonstrate the value of the presented synthetic image generation framework: the new model achieves 23\% better accuracy and improves mean IoU by 0.9.

The Texas Spacecraft Laboratory has used this pipeline with great success over the past year. Developing an object detection model for the NASA Gateway, including synthetic data generation, training, and hyperparameter tuning, was accomplished in just two weeks, demonstrating the pipeline's applicability across target objects. The ability to rapidly define and explore new model architectures within a common framework has proved invaluable for complex problems like estimation, and more team members now contribute regularly to model development thanks to the simplicity of the Starfish, ravenML, and cometML interfaces.

Source code for Starfish\footnote{\url{https://github.com/autognc/starfish}}, sample Starfish scripts\footnote{\url{https://github.com/autognc/starfish-scripts}}, ravenML source code\footnote{\url{https://github.com/autognc/ravenML}}, our ravenML dataset\footnote{\url{https://github.com/autognc/ravenML-dataset-plugins}} and model training\footnote{\url{https://github.com/autognc/ravenML-train-plugins}} plugins, and bran\footnote{\url{https://github.com/autognc/bran}} are available on Github. 

%% file: main.bbl
\begin{thebibliography}{10}
\providecommand{\url}[1]{#1}
\csname url@samestyle\endcsname
\providecommand{\newblock}{\relax}
\providecommand{\bibinfo}[2]{#2}
\providecommand{\BIBentrySTDinterwordspacing}{\spaceskip=0pt\relax}
\providecommand{\BIBentryALTinterwordstretchfactor}{4}
\providecommand{\BIBentryALTinterwordspacing}{\spaceskip=\fontdimen2\font plus
\BIBentryALTinterwordstretchfactor\fontdimen3\font minus
  \fontdimen4\font\relax}
\providecommand{\BIBforeignlanguage}[2]{{%
\expandafter\ifx\csname l@#1\endcsname\relax
\typeout{** WARNING: IEEEtran.bst: No hyphenation pattern has been}%
\typeout{** loaded for the language `#1'. Using the pattern for}%
\typeout{** the default language instead.}%
\else
\language=\csname l@#1\endcsname
\fi
#2}}
\providecommand{\BIBdecl}{\relax}
\BIBdecl

\bibitem{Cassinis2020}
\BIBentryALTinterwordspacing
L.~P. Cassinis, R.~Fonod, E.~Gill, I.~Ahrns, and J.~G. Fernandez,
  \emph{CNN-Based Pose Estimation System for Close-Proximity Operations Around
  Uncooperative Spacecraft}, 2020. [Online]. Available:
  \url{https://arc.aiaa.org/doi/abs/10.2514/6.2020-1457}
\BIBentrySTDinterwordspacing

\bibitem{Park2019}
T.~H. Park, S.~Sharma, and S.~D'Amico, ``Towards robust learning-based pose
  estimation of noncooperative spacecraft,'' 2019.

\bibitem{Proenca2019}
P.~F. Proença and Y.~Gao, ``Deep learning for spacecraft pose estimation from
  photorealistic rendering,'' \emph{2020 IEEE International Conference on
  Robotics and Automation (ICRA)}, pp. 6007--6013, 2020.

\bibitem{Sharma2019}
S.~{Sharma}, C.~{Beierle}, and S.~{D'Amico}, ``Pose estimation for
  non-cooperative spacecraft rendezvous using convolutional neural networks,''
  in \emph{2018 IEEE Aerospace Conference}, 2018, pp. 1--12.

\bibitem{Richardson2016}
E.~{Richardson}, M.~{Sela}, and R.~{Kimmel}, ``3d face reconstruction by
  learning from synthetic data,'' in \emph{2016 Fourth International Conference
  on 3D Vision (3DV)}, 2016, pp. 460--469.

\bibitem{Hinterstoisser2019}
S.~Hinterstoisser, V.~Lepetit, P.~Wohlhart, and K.~Konolige, ``On pre-trained
  image features and synthetic images for deep learning,'' in \emph{Computer
  Vision -- ECCV 2018 Workshops}, L.~Leal-Taix{\'e} and S.~Roth, Eds.\hskip 1em
  plus 0.5em minus 0.4em\relax Cham: Springer International Publishing, 2019,
  pp. 682--697.

\bibitem{camera_pose}
D.~Acharya, K.~Khoshelham, and S.~Winter, ``Bim-posenet: Indoor camera
  localisation using a 3d indoor model and deep learning from synthetic
  images,'' \emph{ISPRS Journal of Photogrammetry and Remote Sensing}, pp.
  245--258, 2019.

\bibitem{Tremblay2018}
J.~{Tremblay}, A.~{Prakash}, D.~{Acuna}, M.~{Brophy}, V.~{Jampani}, C.~{Anil},
  T.~{To}, E.~{Cameracci}, S.~{Boochoon}, and S.~{Birchfield}, ``Training deep
  networks with synthetic data: Bridging the reality gap by domain
  randomization,'' in \emph{2018 IEEE/CVF Conference on Computer Vision and
  Pattern Recognition Workshops (CVPRW)}, 2018, pp. 1082--10\,828.

\bibitem{SeekerSciTech}
\BIBentryALTinterwordspacing
N.~Dhamani, G.~Martin, C.~Schubert, P.~Singh, N.~Hatten, and M.~R. Akella,
  \emph{Applications of Machine Learning and Monocular Vision for Autonomous
  On-Orbit Proximity Operations}.\hskip 1em plus 0.5em minus 0.4em\relax AIAA,
  2020. [Online]. Available:
  \url{https://arc.aiaa.org/doi/abs/10.2514/6.2020-1376}
\BIBentrySTDinterwordspacing

\bibitem{Talwar2020}
D.~{Talwar}, S.~{Guruswamy}, N.~{Ravipati}, and M.~{Eirinaki}, ``Evaluating
  validity of synthetic data in perception tasks for autonomous vehicles,'' in
  \emph{2020 IEEE International Conference On Artificial Intelligence Testing
  (AITest)}, 2020, pp. 73--80.

\bibitem{Seeker1}
B. Banker and S. Askew. 2019. "Seeker 1.0: Prototype Robotic Free Flying
  Inspector Mission Overview," \textit{Proceedings of the Small Satellite
  Conference,} Session XI: Year in Review II, SSC19-X1-04.
  https://digitalcommons.usu.edu/smallsat/2019/all2019/151/.

\bibitem{skr1_blender_sc_models}
M.~Crawford, ``Blender-spaceflight-models: Blender spaceflight models,''
  \url{https://github.com/brickmack/Blender-Spaceflight-models}, 2018, last
  accessed 22 June 2018.

\bibitem{googleAI}
\BIBentryALTinterwordspacing
{Google, LLC}. (2020) Ai platform. [Online]. Available:
  \url{https://cloud.google.com/ai-platform}
\BIBentrySTDinterwordspacing

\bibitem{azureML}
\BIBentryALTinterwordspacing
Microsoft. (2020) Microsoft azure machine learning. [Online]. Available:
  \url{https://azure.microsoft.com/en-us/services/machine-learning/}
\BIBentrySTDinterwordspacing

\bibitem{sagemaker}
\BIBentryALTinterwordspacing
{Amazon Web Services}. (2020) Aws sagemaker. [Online]. Available:
  \url{https://aws.amazon.com/sagemaker/}
\BIBentrySTDinterwordspacing

\bibitem{ec2}
\BIBentryALTinterwordspacing
------. (2020) Aws ec2. [Online]. Available: \url{https://aws.amazon.com/ec2/}
\BIBentrySTDinterwordspacing

\bibitem{sagemaker_pricing}
\BIBentryALTinterwordspacing
------. (2020) Aws sagemaker ml instance pricing. [Online]. Available:
  \url{https://aws.amazon.com/sagemaker/pricing/}
\BIBentrySTDinterwordspacing

\bibitem{ec2_pricing}
\BIBentryALTinterwordspacing
------. (2020) Aws ec2 on-demand pricing for us-east (ohio) region. [Online].
  Available: \url{https://aws.amazon.com/ec2/pricing/on-demand/}
\BIBentrySTDinterwordspacing

\bibitem{paperspace_gradient}
\BIBentryALTinterwordspacing
Paperspace. (2020) Paperspace gradient. [Online]. Available:
  \url{https://gradient.paperspace.com/}
\BIBentrySTDinterwordspacing

\bibitem{allegro_enterprise}
\BIBentryALTinterwordspacing
{allegro.ai}. (2020) allegro.ai enterprise. [Online]. Available:
  \url{https://allegro.ai/enterprise/}
\BIBentrySTDinterwordspacing

\bibitem{polyaxon_enterprise}
\BIBentryALTinterwordspacing
Polyaxon. (2020) Polyaxon gradient. [Online]. Available:
  \url{https://polyaxon.com/polyaxon-ee/}
\BIBentrySTDinterwordspacing

\bibitem{gradient_pricing}
\BIBentryALTinterwordspacing
Paperspace. (2020) Paperspace gradient managed instance pricing. [Online].
  Available:
  \url{https://docs.paperspace.com/gradient/instances/instance-types}
\BIBentrySTDinterwordspacing

\bibitem{MLFlow}
\BIBentryALTinterwordspacing
MLFlow. (2020) Mlflow. [Online]. Available: \url{https://mlflow.org/}
\BIBentrySTDinterwordspacing

\bibitem{Kubeflow}
\BIBentryALTinterwordspacing
Kubeflow. (2020) Kubeflow. [Online]. Available: \url{https://www.kubeflow.org/}
\BIBentrySTDinterwordspacing

\bibitem{allegro_open_source}
\BIBentryALTinterwordspacing
{allegro.ai}. (2020) allegro.ai trains open-source. [Online]. Available:
  \url{https://allegro.ai/trains-open-source/}
\BIBentrySTDinterwordspacing

\bibitem{polyaxon_ce}
\BIBentryALTinterwordspacing
Polyaxon. (2020) Polyaxon community edition. [Online]. Available:
  \url{https://polyaxon.com/polyaxon-ce/}
\BIBentrySTDinterwordspacing

\bibitem{kubernetes}
\BIBentryALTinterwordspacing
Kubernetes. (2020) Kubernetes. [Online]. Available:
  \url{https://kubernetes.io/}
\BIBentrySTDinterwordspacing

\bibitem{DomainRandomization}
J.~{Tobin}, R.~{Fong}, A.~{Ray}, J.~{Schneider}, W.~{Zaremba}, and P.~{Abbeel},
  ``Domain randomization for transferring deep neural networks from simulation
  to the real world,'' in \emph{2017 IEEE/RSJ International Conference on
  Intelligent Robots and Systems (IROS)}, 2017, pp. 23--30.

\bibitem{cometML}
\BIBentryALTinterwordspacing
cometML. (2020) cometml. [Online]. Available: \url{https://www.comet.ml/site/}
\BIBentrySTDinterwordspacing

\bibitem{aws}
\BIBentryALTinterwordspacing
{Amazon Web Services}. (2020) Aws homepage. [Online]. Available:
  \url{https://aws.amazon.com/}
\BIBentrySTDinterwordspacing

\bibitem{aws_gov}
\BIBentryALTinterwordspacing
------. (2020) Aws govcloud us. US-East (Ohio) region used. [Online].
  Available: \url{https://aws.amazon.com/govcloud-us/}
\BIBentrySTDinterwordspacing

\bibitem{neptune.ai}
\BIBentryALTinterwordspacing
neptune.ai. (2020) neptune.ai. [Online]. Available: \url{https://neptune.ai/}
\BIBentrySTDinterwordspacing

\bibitem{Huang_2017}
\BIBentryALTinterwordspacing
J.~Huang, V.~Rathod, C.~Sun, M.~Zhu, A.~Korattikara, A.~Fathi, I.~Fischer,
  Z.~Wojna, Y.~Song, S.~Guadarrama, and et~al., ``Speed/accuracy trade-offs for
  modern convolutional object detectors,'' \emph{2017 IEEE Conference on
  Computer Vision and Pattern Recognition (CVPR)}, Jul 2017. [Online].
  Available: \url{http://dx.doi.org/10.1109/CVPR.2017.351}
\BIBentrySTDinterwordspacing

\end{thebibliography}
